\documentclass[entropy,article,submit,moreauthors,pdftex]{Definitions/mdpi} 
\usepackage{float}
\usepackage{amsmath}
\usepackage{amsfonts}
\usepackage{mathtools}
\usepackage{amssymb}
\usepackage{float}
\usepackage{xcolor}
\usepackage{enumerate}
\usepackage{isomath}
\usepackage{bm}
\usepackage{cite}
\usepackage{caption}
\usepackage{subfigure}
\usepackage{graphicx}
\usepackage{algorithm,algcompatible,amsmath}
\algnewcommand\INPUT{\item[\textbf{Input:}]}%
\algnewcommand\OUTPUT{\item[\textbf{Output:}]}%
\restylefloat{table}
\usepackage{tikz} 
\usepackage{lipsum}
\usetikzlibrary{arrows, positioning, automata}
\usetikzlibrary{shapes.geometric, arrows}
\def\xfoo#1^#2\relax#3\valign{%
\mathbf{#1}\ifx\valign#2\valign\else^{\mathbf{#2}}\fi}

\def\Lm{{\mathcal{L}}}

\def\Ber{{\rm Bernoulli}}

\def\jnt{{\rm joint}}

\def \Xt{\widetilde{X}}
\def \Yt{\widetilde{Y}}

\def\mset{{Z^m_{1:N}}}

\def\trte{{\rm sep}}

\def\Ztrain{Z^{\rm m_{tr}}}
\def\Ztest{Z^{\rm m_{te}}}
\def\Ztr{D^{\rm m_{tr}}}
\def\mtr{m_{\rm tr}}
\def\Zte{D^{\rm m_{te}}}
\def\mte{m_{\rm te}}

\def\Yt{\widetilde{Y}}

\def \Nscr{\mathcal{N}}
\def \Wscr{\mathscr{W}}

\def \DL{ { \Delta} L}
\def \DLm{ {\Delta} {\mathcal{ L}}}

\usepackage{amsmath, amssymb, bbm, xspace}
\usepackage{epsfig}
\usepackage{longtable}
\usepackage{color}
\usepackage{mathrsfs}
\usepackage{comment}

\usepackage{courier}

\def\bkE{{\rm I\kern-.17em E}}
\def\bk1{{\rm 1\kern-.17em l}}
\def\bkD{{\rm I\kern-.17em D}}
\def\bkR{{\rm I\kern-.17em R}}
\def\bkP{{\rm I\kern-.17em P}}

\def\bkZ{{\bf{Z}}}

\def\bkE{{\rm I\kern-.17em E}}
\def\bk1{{\rm 1\kern-.17em l}}
\def\bkD{{\rm I\kern-.17em D}}
\def\bkR{{\rm I\kern-.17em R}}
\def\bkP{{\rm I\kern-.17em P}}

\makeatletter
\newcommand{\pushright}[1]{\ifmeasuring@#1\else\omit\hfill$\displaystyle#1$\fi\ignorespaces}
\newcommand{\pushleft}[1]{\ifmeasuring@#1\else\omit$\displaystyle#1$\hfill\fi\ignorespaces}
\makeatother

\def\bkZ{{\bf{Z}}}
\def\b12{(\beta_1,\beta_2)}

\renewcommand{\theexample}{\thesection.\arabic{example}}

\renewcommand{\theremark}{\thesection.\arabic{remark}}

\def\Xscr{\mathcal{X}}
\def\Yscr{\mathcal{Y}}

\def\Ebb{\mathbb{E}}
\newlength{\noteWidth}
\long\def\notes#1{\ifinner
{\tiny #1}
\else
\marginpar{\parbox[t]{\noteWidth}{\raggedright\tiny #1}}
\fi\typeout{#1}}

 \def\notes#1{\typeout{read notes: #1}} 

\newcommand{\ie}{i.e.\@\xspace} 
\newcommand{\eg}{e.g.\@\xspace} 

\newcommand{\Real}{\ensuremath{\mathbb{R}}}

\def\Ebb{\mathbb{E}}

\def\exp{\mathop{\hbox{\rm exp}}}

\def\etal{{\em et al.\ }}  

\def\spose#1{\hbox to 0pt{#1\hss}}

\def\text #1{\hbox{\quad#1\quad}}

\def\nthinsp{\mskip -2   mu}

\def\P{_{\scriptscriptstyle P}}

\def\S{_{\scriptscriptstyle S}}

\def\superstar{^{\raise 0.5pt\hbox{$\nthinsp *$}}}
\def\SUPERSTAR{^{\raise 0.5pt\hbox{$*$}}}

\def\lamstarT {\lambda^{\raise 0.5pt\hbox{$\nthinsp *$}T}}

\def\Ascr{{\cal A}}

\def\Lscr{{\cal L}}

\def\Pscr{{\cal P}}

\def\Uscr{{\cal U}}

\def\Wscr{{\cal W}}

\def\Nscr{{\cal N}}

\def\Zscr{{\cal Z}}
\def\Xscr{{\cal X}}
\def\Yscr{{\cal Y}}

\def\non{\nonumber}

\let\forallnew\forall
\renewcommand{\forall}{\forallnew\ }
\let\forall\forallnew

		\def\bkE{{\rm I\kern-.17em E}}
		\def\bk1{{\rm 1\kern-.17em l}}
		\def\bkD{{\rm I\kern-.17em D}}
		\def\bkR{{\rm I\kern-.17em R}}
		\def\bkP{{\rm I\kern-.17em P}}
		\def\bkY{{\bf \kern-.17em Y}}
		\def\bkZ{{\bf \kern-.17em Z}}
		\def\bkC{{\bf  \kern-.17em C}}

{\begin{list}{}%
         {\setlength{\leftmargin}{#1}}%
         \item[]%
}
{\end{list}}

		\def\bsp{\begin{split}}
		\def\beq{\begin{eqnarray}}
		\def\bal{\begin{align*}}
		\def\bc{\begin{center}}
		\def\be{\begin{enumerate}}
		\def\bi{\begin{itemize}}
		\def\bs{\begin{small}}
		\def\bS{\begin{slide}}
		\def\ec{\end{center}}
		\def\ee{\end{enumerate}}
		\def\ei{\end{itemize}}
		\def\es{\end{small}}
		\def\eS{\end{slide}}
		\def\eeq{\end{eqnarray}}
		\def\eal{\end{align*}}
		\def\esp{\end{split}}
		\def\qed{ \vrule height7.5pt width7.5pt depth0pt}  

	\def\cp2problem#1#2#3#4{\fbox
		 {\begin{tabular*}{0.9\textwidth}
			{@{}l@{\extracolsep{\fill}}l@{\extracolsep{6pt}}l@{\extracolsep{\fill}}c@{}}
				#1 & & $#4 $ 
			\end{tabular*}}}

		\def\bkE{{\rm I\kern-.17em E}}
		\def\bk1{{\rm 1\kern-.17em l}}
		\def\bkD{{\rm I\kern-.17em D}}
		\def\bkR{{\rm I\kern-.17em R}}
		\def\bkP{{\rm I\kern-.17em P}}
		
		\def\bkZ{{\bf{Z}}}

\newcommand {\beeq}[1]{\begin{equation}\label{#1}}
\newcommand {\eeeq}{\end{equation}}
\newcommand {\bea}{\begin{eqnarray}}
\newcommand {\eea}{\end{eqnarray}}

\def\texitem#1{\par\smallskip\noindent\hangindent 25pt
               \hbox to 25pt {\hss #1 ~}\ignorespaces}

\def\bsp{\begin{split}}
		\def\beq{\begin{eqnarray}}
		\def\bal{\begin{align*}}
		\def\bc{\begin{center}}
		\def\be{\begin{enumerate}}
		\def\bi{\begin{itemize}}
		\def\bs{\begin{small}}
		\def\bS{\begin{slide}}
		\def\ec{\end{center}}
		\def\ee{\end{enumerate}}
		\def\ei{\end{itemize}}
		\def\es{\end{small}}
		\def\eS{\end{slide}}
		\def\eeq{\end{eqnarray}}
		\def\eal{\end{align*}}
		\def\esp{\end{split}}
		\def\qed{ \vrule height7.5pt width7.5pt depth0pt}  

\usepackage{amsmath, amssymb, xspace}
\usepackage{epsfig}
\usepackage{longtable}
\usepackage{color}
\usepackage{mathrsfs}

\def\Nscr{{\cal N}}

\pdfoutput=1
\preto{\abstractkeywords}{\nolinenumbers}

\Title{INFORMATION-THEORETIC GENERALIZATION BOUNDS FOR META-LEARNING AND APPLICATIONS}

\Author{Sharu Theresa Jose $^{1,}$* and Osvaldo Simeone $^{2}$}

\AuthorNames{Sharu Theresa Jose and Osvaldo Simeone}

\address{%
$^{1}$ \quad Postdoctoral Researcher, Department of Engineering, King's College London; sharu.jose@kcl.ac.uk\\
$^{2}$ \quad Professor, Department of Informatics, King's College London; osvaldo.simeone@kcl.ac.uk}

\corres{Correspondence: sharu.jose@kcl.ac.uk}

\abstract{Meta-learning, or ``learning to learn'', refers to techniques that infer an inductive bias from data corresponding to multiple related tasks with the goal of improving the sample efficiency for new, previously unobserved, tasks. A key performance measure for meta-learning is the meta-generalization gap, that is, the difference between the average loss measured on the meta-training data and on a new, randomly selected task. This paper presents novel information-theoretic upper bounds on the meta-generalization gap. Two broad classes of meta-learning algorithms are considered that uses either separate within-task training and test sets, like MAML, or joint within-task training and test sets, like Reptile. Extending the existing work for conventional learning, an upper bound on the meta-generalization gap is derived for the former class that depends on the mutual information (MI) between the output of the meta-learning algorithm and its input meta-training data. For the latter, the derived bound includes an additional MI between the output of the per-task learning procedure and corresponding data set to capture within-task uncertainty. Tighter bounds are then developed for the two classes via novel Individual Task MI (ITMI) bounds. Applications of the derived bounds are finally discussed, including a broad class of noisy iterative algorithms for meta-learning.}


\begin{document}


\section{Introduction}
As formalized by the ``no free lunch theorem'', 
any effective learning procedure must be based on  prior assumptions on the task of interest \citep{shalev2014understanding}. These include the selection of a model class and of the hyperparameters of a learning algorithm, such as weight initialization and learning rate. In conventional single-task learning, these assumptions, collectively known as \textit{inductive bias}, are fixed \textit{a priori} relying on domain knowledge or validation \citep{shalev2014understanding,bishop2006pattern,simeone2018brief}. Fixing a suitable inductive bias can significantly reduce the sample complexity of the learning process, and is thus crucial to any learning procedure. The goal of \textit{meta-learning} is to automatically infer the inductive bias, thereby  \textit{learning to learn} from past experiences via the observation of a number of related tasks, so as to speed up learning a new and unseen task \citep{schmidhuber1987evolutionary, thrun1998learning, thrun1996learning, vilalta2002perspective,simeone2020learning}.

In this work, we consider the meta-learning problem of inferring the hyperparameters of a learning algorithm.  The learning algorithm (henceforth, called base-learning algorithm or base-learner) is defined as a stochastic mapping $P_{W|Z^m,u}$ from the input training set $Z^m=(Z_1,\hdots,Z_m)$ of $m$ samples to a model parameter $W \in \Wscr$ for a fixed hyperparameter vector $u$. The meta-learning algorithm (or meta-learner) infers the hyperparameter vector $u$, which defines the inductive bias, by observing a finite number of related tasks.

For example, consider the well-studied algorithm of \textit{biased regularization} for supervised learning \citep{kuzborskij2017fast,kienzle2006personalized}. Let us denote each data point $Z=(X,Y)$ as a tuple of input features $X \in \Real^d$ and label $Y \in \Real$. The loss function $l: \Wscr \times \Zscr \rightarrow \Real$ is given as the quadratic measure $l(w,z)=(\langle w,x \rangle -y)^2$ that quantifies the loss accrued by the inferred model parameter $w$ on a data sample $z$. Corresponding to each per-task data set $Z^m$, the biased regularization algorithm $P_{W|Z^m,u}$ is a Kronecker delta function centered at the minimizer of the following optimization problem
\begin{align}
    \frac{1}{m}\sum_{j=1}^m l(w,Z_j) + \frac{\lambda}{2} ||w-u||^2,
\end{align}
which corresponds to an empirical risk minimization problem with a biased regularizer. Here,
$\lambda>0$ is a regularization constant that weighs the deviation of the model parameter $w$  from a bias vector $u$. The bias vector $u$ can be then thought of as a common ``mean'' among related tasks. In the context of meta-learning, the objective then is to infer the bias vector $u$ by observing  data sets from a number of similar related tasks. Different meta-learning algorithms have been developed for this problem \citep{denevi2019learning,denevi2020advantage}.

In the meta-learning problem under study, we follow the standard setting of  Baxter \citep{baxter2000model} and assume that the learning tasks belong to a \textit{task environment}, which is defined by a probability distribution $P_T$ on the space  of  learning tasks $\mathcal{T}$, and per-task data distributions $\{P_{Z|T=\tau}\}_{\tau \in \mathcal{T}}$. The data set $Z^m$ for a task $\tau$ is then generated i.i.d. according to the distribution $P_{Z|T=\tau}$.  The meta-learner observes the performance of the base-learner on the \textit{meta-training data} from a finite number of \textit{meta-training tasks}, which are sampled independently from the task environment, and infers the hyperparameter $U$ such that it can learn a new task,  drawn from the same task environment, from fewer data samples.
 
The quality of the inferred hyperparameter $U$ is measured by the \textit{meta-generalization loss}, $\Lm_{g}(U)$,  which is the average loss incurred on the data set $Z^m \sim P_{Z^m|T}$ of a new, previously unseen task $T$ sampled from the task distribution $P_T$. The notation will be formally introduced in Section~\ref{sec:gap_fullsupport}.  While the goal of meta-learning is to infer a hyperparameter $U$ that minimizes the meta-generalization loss $\Lm_{g}(U)$, this is not computable, since the underlying task and data distributions are unknown. Instead, the meta-learner can evaluate an empirical estimate of the loss, $\Lm_{t}(U|\mset)$, using the meta-training set $\mset$ of data from $N$ tasks, which is referred to as \textit{meta-training loss}. The difference between the meta-generalization loss and the meta-training loss is the \textit{meta-generalization gap},
\begin{align}
\Delta \Lm(U|\mset)=\Lm_{g}(U)-\Lm_{t}(U|\mset), \label{eq:testloss_decomposition}
    \end{align}
and measures how well the inferred hyperparameter $U$ generalizes to a new, previously unseen task. In particular, if the meta-generalization gap is small, on average or with high probability, then the performance of the meta-learner on the meta-training set can be taken as a reliable estimate of the meta-generalization loss.

  
In this paper, we study information-theoretic upper bounds on the \textit{average meta-generalization gap } $\Ebb_{P_{\mset}P_{U|\mset}}[\Delta \Lm(U|\mset)]$, where the average is with respect to the meta-training set $\mset$ and the meta-learner defined by the stochastic kernel $P_{U|\mset}$. Specifically, we extend the recent line of work initiated by Russo and Zhou \citep{russo2016controlling}, and Xu and Raginsky \citep{xu2016learning} which obtain mutual information (MI)-based bounds on the average generalization gap for conventional learning, to meta-learning. To the best of our knowledge, this is the first work that studies information-theoretic bounds for meta-learning. 

The bounds on average meta-generalization gap, studied in this work, are distinct from the other well-known bounds on meta-generalization gap in literature. Broadly speaking,  
existing bounds on the meta-generalization gap can be grouped into two - high probability probably-approximately-correct (PAC) bounds, and high probability PAC-Bayesian bounds. These upper bounds take the general form, $\Ebb_{P_{U|\mset}}[\Delta \Lscr(U|\mset)] \leq \epsilon$, that hold with probability at least $1-\delta$, for $\delta \in (0,1)$, over the meta-training set $\mset$. In contrast, our work focuses on bounding $\Ebb_{P_{\mset}}\Ebb_{P_{U|\mset}}[\Delta \Lscr(U|\mset)]$ on average also over the meta-training set. Notable PAC bounds on meta-generalization gap include the bound of  Baxter \citep{baxter2000model} obtained using the framework of Vapnik-Chervonenkis (VC) dimensions; and of Maurer \citep{maurer2005algorithmic} which employs the algorithmic stability \citep{devroye1979distribution, rogers1978finite} properties. In contrast, the PAC-Bayesian bounds also incorporate prior beliefs on the base-learner and the meta-learner posteriors via an auxiliary data-independent prior distribution $Q_{W|U}$ and a hyper-prior distribution $Q_U$, respectively. Most notable PAC-Bayesian bounds include that of  Pentina and Lambert \citep{pentina2014pac}, the tighter bound of Amit and Meir \citep{ amit2018meta}, and most recently the bounds of Rothfuss \etal \citep{rothfuss2020pacoh}. 
 While the high-probability bounds are agnostic to task and data distributions, our information-theoretic bounds depend explicitly on the task and per-task data distributions, on the loss function, and on the meta-training algorithm, in accordance to prior work on information-theoretic generalization bounds.
 
 Another general property inherited from the information-theoretic approach adopted in this paper is that the bounds on the average meta-generalization gap under study are designed to hold for arbitrary base-learners and meta-learners. As such, they generally do not result in tighter bounds as compared to  non-information theoretic generalization guarantees obtained for specific meta-learning problems, such as the ridge regression problem with meta-learned bias vector mentioned above \citep{denevi2018incremental}. In contrast, the general purpose of the bounds in this paper is to provide insights into the number of tasks, and number of samples per task required to ensure that the training based metrics are a good approximation to their population counterparts. 
\subsection{Main Contributions}

The derivation of bounds on average
meta-generalization gap differs from conventional learning owing to two levels of uncertainties -- \textit{environment-level} uncertainty and \textit{within-task} uncertainty. While within-task uncertainty results from observing a finite number $m$ of data samples per task as in  conventional learning, environment-level uncertainty results from observing a finite number $N$ of tasks from the task-environment. The relative importance of these two forms of uncertainty depend on the use made by the meta-learner of the meta-training data.
In fact, depending on how the meta-training data is used by the meta-learner, we identify two main classes of meta-training algorithms -- with \textit{separate within-task training and test sets}, and \textit{joint within-task training and test sets}. 
The former class includes the state-of-the-art meta-learning algorithms such as Model Agnostic Meta-Learning (MAML) \citep{finn2017model} that split the training data corresponding to each task into training and test sets, with the latter reserved for within-task validation. In contrast, the second class of algorithms, such as Reptile \citep{nichol2018first}, use the entire per-task data both for training and testing. Our main contributions are as follows.\\
\noindent $\bullet$ In Theorem~\ref{thm:FMMIbound_different}, we show that, for the case with \textit{separate} within-task training and test sets, the average meta-generalization gap contains only the contribution of environment-level uncertainty. This is captured by a ratio of the mutual information (MI) between the output of the meta-learner $U$ and the meta-training set $\mset$, and the number of tasks $N$ as
\begin{align}
\biggl|\Ebb_{P_{\mset}P_{U|\mset}}\bigl[\DLm^{
\trte} (U|\mset)\bigr]\biggr|&\leq \sqrt{\frac{2 \sigma^2}{N} I(U;\mset)}, \label{eq:MIboundsep_intr}
\end{align}
where $\sigma^2$ is the sub-Gaussianity variance factor of the meta-loss function. This
is a direct parallel of the MI-based bounds for single-task learning \citep{xu2017information}.\\
\noindent $\bullet$ In Theorem~\ref{thm:FMMIbound_same}, we then shown that, for the case with \textit{joint} within-task training and test sets, the bound on the average meta-generalization gap also contains a contribution due to the within-task uncertainty via the ratio of the MI between the output of the base-learner and within task training data and the per-task data sample size $m$. Specifically, we have the following bound
\begin{align}
\biggl| \Ebb_{P_{\mset}P_{U|\mset}}[\DLm^{\jnt}(U|\mset)]\biggr|&\leq
\sqrt{\frac{2 \sigma^2}{N}I(U;\mset)}+\Ebb_{P_T}\biggl[ \sqrt{\frac{2 \delta_T^2}{m} I(W;Z^m|T=\tau)}\biggr], \label{eq:MIboundjnt_intr}
\end{align} where $\delta_T^2$ is the sub-Gaussianity variance factor of the loss function $l(w,z)$ for task $T$.
\\
\noindent $\bullet$ In Theorem~\ref{thm:ITMI_separate} and Theorem~\ref{thm:ITMIbound_same}, we extend the individual sample MI (ISMI) bound of \citep{bu2019tightening} to obtain novel Individual Task MI (ITMI)-based bounds on the meta-generalization gap for both separate and within-task training and test sets as
\begin{align}
\biggl|\Ebb_{{P_{\mset}P_{U|\mset}}}\bigl[\DLm^{
\trte} (U|\mset)\bigr] \biggr|&\leq \frac{1}{N}\sum_{i=1}^N \sqrt{2 \sigma^2 I(U;Z_i^m)},
\end{align} and 
\begin{align}
 &\biggl| \Ebb_{{P_{\mset}P_{U|\mset}}}[\DLm^{\jnt}(U|\mset)]\biggr|
 \leq  \frac{1}{N}\sum_{i=1}^N  
\sqrt{2 \sigma^2 I(U;Z^m_i)} +\Ebb_{P_T}\biggl[ \frac{1}{m}\sum_{j=1}^m\sqrt{2 \delta_{T}^2 I(W;Z_j|T=\tau)}\biggr].
\end{align}
These bounds can be seen to be tighter than the MI-based bounds in \eqref{eq:MIboundsep_intr} and \eqref{eq:MIboundjnt_intr}, respectively.\\
\noindent $\bullet$ Finally, we study the applications of the derived bounds to two meta-learning problems. The first is a parameter estimation setup that involves one-shot meta-learning and base-learning procedures, for which a closed form expression for meta-generalization gap can be derived. The second application covers a broad range of noisy iterative meta-learning algorithms and is inspired by the work of Pensia \etal \citep{pensia2018generalization} for conventional learning.
 

\subsection{Related Work}
For conventional learning, there exists a rich literature on diverse frameworks for deriving upper bounds on the generalization gap, \ie on the difference between generalization and training losses. Classical bounds from statistical learning theory quantify the generalization gap in terms of measures of complexity of the model class, most notably  VC dimension \citep{vapnik2015uniform} and Radmacher complexity \citep{koltchinskii2000rademacher}. This approach obtains high-probability probably approximate correct (PAC) bounds on the generalization gap with respect to the training set. An alternate line of high-probability bounding techniques relies on the notion of 
\textit{algorithmic stability}, which measures the sensitivity of the output of a learning algorithm to the replacement of individual samples from the training data set. The pioneering work \citep{bousquet2002stability} has been extended to include various notions of algorithmic stability \citep{kearns1999algorithmic, poggio2004general,kutin02almost}. As a notable example, a distributional notion of stability in terms of \textit{differential privacy}, which quantifies the sensitvity of the distribution of algorithm's output to data set, has been studied in \citep{dwork2015preserving, bassily2016algorithmic}. The high-probability PAC-Bayesian bounds rely on change of measure arguments and uses the Kullback Leibler (KL) divergence between the algorithm and a data-independent prior to quantify the algorithmic sensitivity \citep{mcallester1999pac, seeger2002pac, alquier2016properties}.

Following the initial work of Russo and Zou \citep{russo2016controlling}, information-theoretic bounds on the \textit{average generalization gap} for conventional learning have been widely investigated in recent years. Xu and Raginsky \citep{xu2017information} showed that the MI between the output of the learning algorithm and its training data set yields an upper bound bound in expectation on the generalization gap. The bound has been shown to offer computable generalization gaurentees for noisy iterative algorithms including Stochastic Gradient Langevin Dynamics (SGLD) in \citep{pensia2018generalization}. Various refinements of the MI-based bound have since been analyzed to obtain tighter bounds. In particular, the bounds in \citep{asadi2018chaining} employ chaining mutual information techniques to tighten the bounds in \citep{xu2017information}, while the bound in \citep{bu2019tightening} depend on the MI between the output of the algorithm and an individual data sample.  The MI between the  output of the algorithm and a random subset of the data set appears in the bounds introduced in \citep{negrea2019information}. 
The total variation information between the joint distribution of the training data and algorithmic output and the product of marginals was shown in \citep{alabdulmohsin2020towards} to yield a bound on the generalization gap  for any bounded loss function. Subsequent works in \citep{jiao2017dependence, issa2018computable, issa2019strengthened} consider other information-theoretic measures, such as maximum leakage and lautum information. Most recently, a conditional mutual information (CMI)-based approach has been proposed in  \citep{steinke2020reasoning} to develop generalization bounds.

\subsection{Notation}
Throughout this paper, upper case letters, \eg $X$, denote random variables and lower case letters, \eg $x$, their realizations.
We use $\Pscr(\cdot)$ to denote the set of all probability distributions on the argument set or vector space.
For a discrete or continuous random variable $X$ taking values in a set or vector space $\Xscr$, $P_X \in \Pscr(\Xscr)$ denotes its probability distribution, with $P_X(x)$ being the probability mass or density value at $x \in \Xscr$. We denote as $P_{X^n}$ the $n$-fold product distribution induced by $P_X$. The conditional distribution of a random variable $X$ given random variable $Y$ is similarly defined as $P_{X|Y}$, with $P_{X|Y}(x|y)$ representing the probability mass or density at $X=x$ conditioned on the event $Y=y$. We use $|| \cdot||_2$ to denote the Euclidean norm of the argument vector, and $I_d$ to denote a $d$-dimensional identity matrix. We define the Kronecker delta $\delta(x-x_0)=1$ if $x=x_0$ and $\delta(x-x_0)=0$ otherwise.
\section{Problem Definition}
In this section, we define the problem of interest by introducing the key definitions of generalization gap for  conventional, or single-task, learning and for meta-learning.
\subsection{Generalization Gap for Single-Task Learning}
Consider first the conventional problem of learning a task $\tau \in \mathcal{T}$.
\begin{figure}[h!]
\centering
\includegraphics[scale=0.28,clip=true, trim = 0.3in  0.1in 0.4in 0.15in]{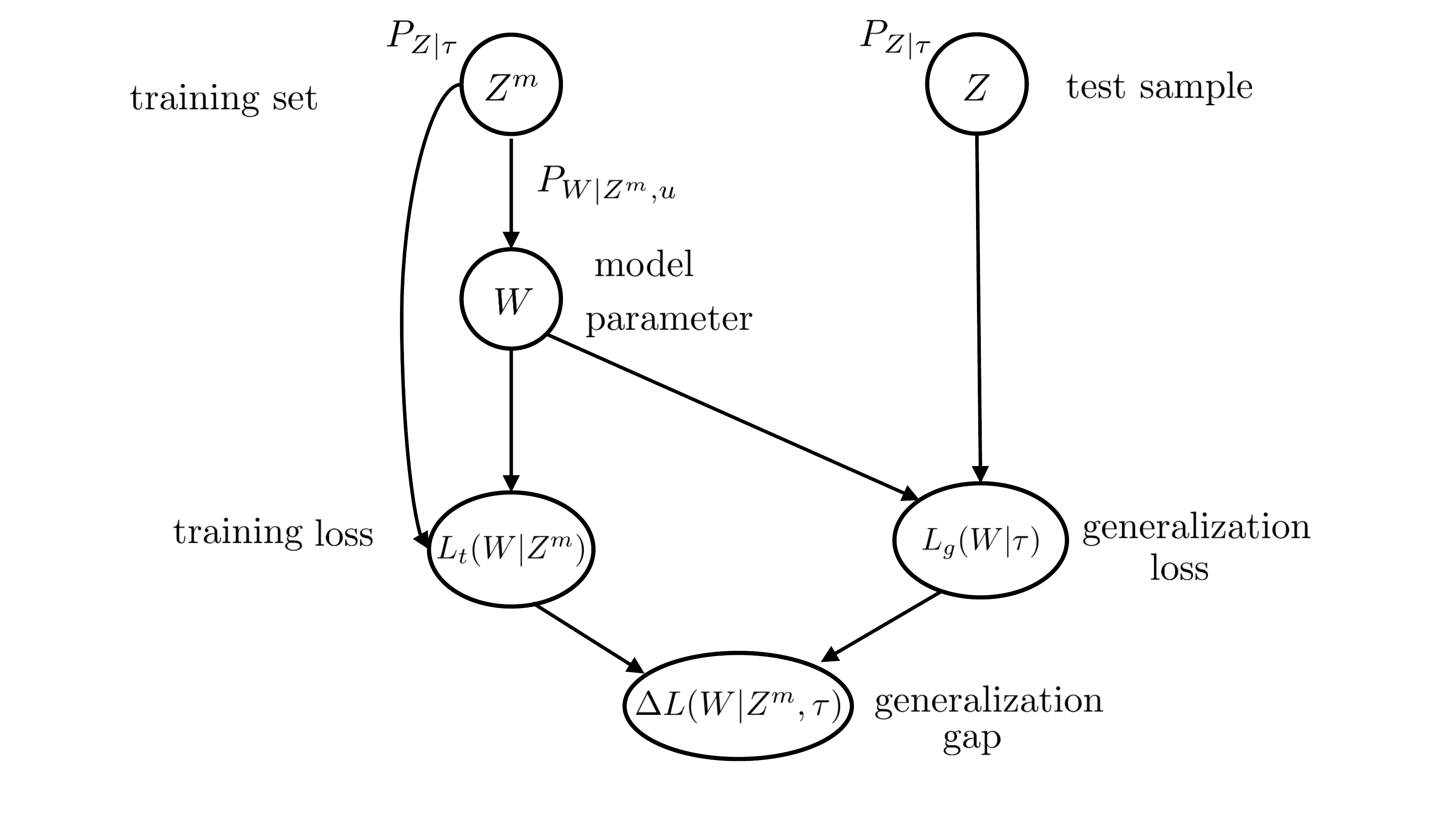} 
\caption{ Directed graph representing the variables involved in the definition of generalization gap \eqref{eq:gap_singletask} for single-task learning.}\label{fig:BN_singletask}
\end{figure}
As illustrated in Figure~\ref{fig:BN_singletask}, each task $\tau \in \mathcal{T}$ 
is associated with an underlying \textit{unknown} data distribution, $P_{Z|T=\tau} \in \Pscr(\Zscr)$, defined in a subset or vector space $\Zscr$. Henceforth, we use $P_{Z|\tau}$ to denote $P_{Z|T=\tau}$ for notational convenience. 
 The training procedure, which is referred to as the \textit{base-learner}, has  access to a training data set $Z^m=(Z_1,Z_2, \hdots,Z_m) \sim P_{Z^m|\tau}$  of $m$ independent and identically distributed (i.i.d.) samples drawn from distribution $P_{Z|\tau}$. The base-learner uses this data set to choose a model, or hypothesis, $W$ from the model class $\Wscr$ by using a \textit{randomized} training procedure defined by a conditional distribution $P_{W|Z^m,u}$ as
 \begin{align}
W \sim P_{W|Z^m,u}. \label{eq:baselearner_training}\end{align}
 The conditional distribution $P_{W|Z^m,u}$ defines a stochastic mapping from the training data set $Z^m$ to the model class $\Wscr$. 
The training procedure  \eqref{eq:baselearner_training} is parameterized by a vector $u \in \Uscr$ of \textit{hyperparameters}, which defines the inductive bias. As an example, the base-learner $P_{W|Z^m,u}$ may follow Stochastic Gradient Descent (SGD) updates with hyperparameters $u$ including the learning rate and the initialization point. 

 The performance of a parameter vector $w \in \Wscr$ on a data sample $z \in \Zscr$ is measured by a loss function $l: \Wscr \times \Zscr \rightarrow \Real_{+}$. 
The \textit{generalization loss} for a model parameter vector $w \in \Wscr$ is the average
\begin{align}
&L_g(w|\tau)=\Ebb_{P_{Z|\tau}}[l(w,Z)], \label{eq:genloss}
\end{align} over a test example $Z$ independently drawn from the data distribution $P_{Z|\tau}$. The subscript $g$ is used to distinguish the generalization loss from the training loss defined below. The generalization loss cannot be computed by the learner, given that the data distribution $P_{Z|\tau}$ is unknown. Instead, the learner can evaluate the  \textit{training loss} on the data set $Z^m$, which is defined as the empirical average
\begin{align}
 L_t(w|Z^m)=\frac{1}{m}\sum_{i=1}^{m}l(w,Z_i). \label{eq:trainingloss}
\end{align} The subscript $t$ specifies that the loss is the empirical training loss.

The difference between generalization loss \eqref{eq:genloss} and training loss \eqref{eq:trainingloss} is known as \textit{generalization gap},
\begin{align}
\Delta L(w|Z^m,\tau)=L_g(w|\tau)-L_t(w|Z^m),
\end{align} and is a key metric that quantifies the level of uncertainty\footnote{This type of uncertainty is known as epistemic.} at the learner regarding the data distribution $P_{Z|\tau}$.
The average generalization gap for the data distribution $P_{Z|\tau}$ and base-learner $P_{W|Z^m,u}$ is defined as
\begin{align}
\Ebb_{P_{Z^m,W|\tau,u}}[\Delta L(W|Z^m,\tau)] , \label{eq:gap_singletask}
\end{align}where the expectation is taken with respect to the joint distribution $P_{Z^m,W|\tau,u}=P_{Z^m|\tau} P_{W|Z^m,u}$. A summary of the variables involved in the Definition of the generalization gap \eqref{eq:gap_singletask} can be found in Figure~\ref{fig:BN_singletask}.
 
 Intuitively, if the generalization gap is small, on average or with high probability, then the base-learner can take the performance \eqref{eq:trainingloss} on the training set $Z^m$ as a reliable measure of the generalization loss \eqref{eq:genloss}  of the trained model $W$. Furthermore, data-dependent bounds on the generalization gap can be used as regularization terms to avoid overfitting, yielding generalized Bayesian inference problems \citep{knoblauch2019generalized}, \citep{bissiri2016general}. 
\subsection{Generalization Gap for Meta-Learning}\label{sec:gap_fullsupport}
As discussed, in single-task learning, the inductive bias $u$, defining the  hyperparameters of the training procedure, must be selected a priori, \ie, without having access to task-specific data. The inductive bias determines the training data set size $m$ needed to ensure a small generalization loss \eqref{eq:genloss}, since, generally speaking, richer models require more data to be trained \citep{shalev2014understanding}. The sample complexity can be generally reduced if one selects a suitable inductive bias based on prior information. Such prior information is typically obtained from domain knowledge on the problem under study. In contrast, meta-learning aims at automatically inferring an effective inductive bias based on data from related tasks.

To elaborate, we follow the setting of \citep{baxter2000model}, in which a meta-learner observes data from a number of tasks, known as \textit{meta-training tasks}, from the same \textit{task environment}.
A task environment is defined by a task distribution $P_T \in \Pscr(\mathcal{T})$, supported on the space $\mathcal{T}$ of tasks, and by a per-task data distribution $P_{Z|\tau}$ for each task $\tau \in \mathcal{T}$. Using the meta-training data drawn from a randomly selected subset of tasks, the meta-learner infers a hyperparameter vector $u \in \Uscr$ defining the inductive bias. This is done with the goal of ensuring that, using hyperparameter $u$, the base-learner $P_{W|Z^m,u}$ can efficiently learn on a new task, referred to as \textit{meta-test task}, drawn independently from the same task distribution $P_T$. 
%

To elaborate, the meta-training data consists of $N$ data sets $\mset=(Z^m_1, \hdots, Z^m_N)$. Each $i$th data set is generated independently by first drawing a task $T_i \sim P_T$ from the task environment and then a task-specific training data set $Z^m_{i}\sim P_{Z^m|T_i}$. The meta-learner uses the meta-training data set $\mset$ to infer a hyperparameter vector $u \in \Uscr$. To this end, we consider a \textit{randomized meta-learner}
\begin{align}
U \sim P_{U|\mset}, \label{eq:meta_learner}
\end{align}
where $P_{U|\mset}$ is a stochastic mapping from the meta-training set $\mset$ to the space $\Uscr$ of hyperparameters. We distinguish two different formulations of meta-learning that are often considered in the literature. In the first, the per-task data set $Z^m$ is split into training, or support, and test, or query subsets \citep{yin2019meta}, \citep{finn2017model}; while, in the second, the entire data set $Z^m$ is used for both within-task training and testing \citep{baxter2000model, pentina2014pac, amit2018meta}.
\subsubsection{Separate Within-Task Training and Test Sets}\label{sec:gap_separatesupport}
\begin{figure}[h!]
\centering
\includegraphics[scale=0.3,clip=true, trim = 0in  0in 0in 0in]{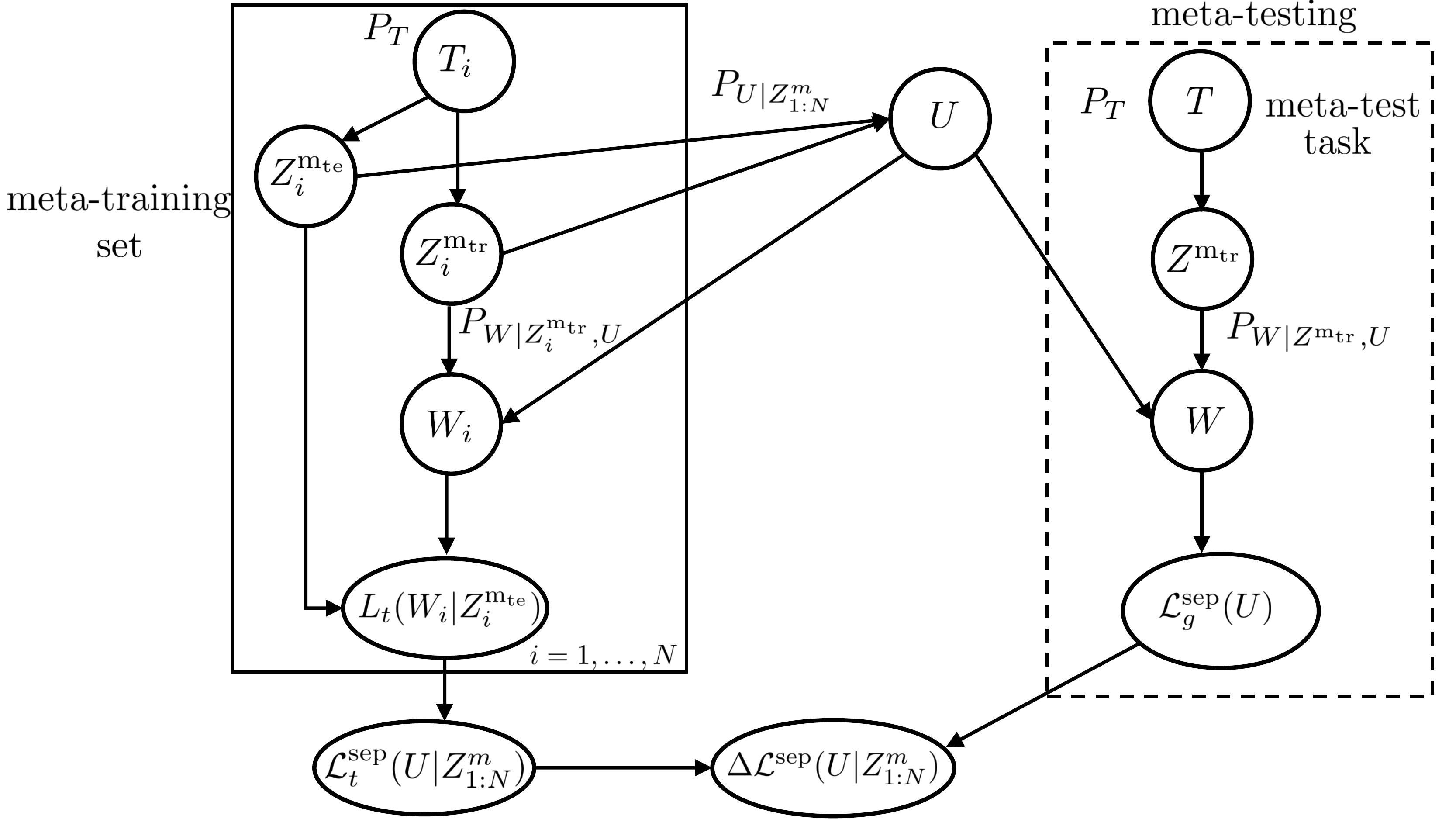} 
\caption{ Directed graph representing the variables involved in the definition of meta-generalization gap \eqref{eq:metagen_gap2} for separate within-task training and testing sets. }\label{fig:BN_separate}
\end{figure}
As seen in Figure~\ref{fig:BN_separate}, in this first approach to meta-learning, each meta-training sub data set $Z^m_i$ is split into a training set and a test set as $Z^m_i=(\Ztrain_i,\Ztest_i)$, where $\Ztrain_i$ contains $\mtr$ i.i.d. training examples  and $\Ztest_i$ contains $\mte$ i.i.d. test examples, with $m=\mtr+\mte$. The \textit{within-task base-learner} $P_{W|\Ztrain_i,u} \in \Pscr(\Wscr)$ maps the per-task training subset $\Ztrain_i$ to random model parameter $W_i \sim P_{W|\Ztrain_i,u}$ for a given hyperparameter $U=u$. The test subset is used to evaluate 
 the empirical  training loss of a model $w$ for task $T_i$ as
\begin{align}
L_{t}(w|\Ztest_i)=\frac{1}{\mte}\sum_{j=1}^{\mte}  l(w, Z^{\mte}_{i,j}),
\end{align}
where $Z^{\mte}_{i,j}$ denote the $j$th example of the test subset $Z^{\mte}_i$.
Furthermore, the overall empirical \textit{meta-training loss} for a hyperparameter $u$ is computed by summing over all meta-training tasks as
\begin{align}
\Lscr_t^{\trte}(u|\mset)=\frac{1}{N}\sum_{i=1}^N L^{\trte}_t(u|Z^m_i), \label{eq:metatrainingloss_sep}
\end{align}
where \begin{align}
L^{\trte}_t(u|Z^m)=\Ebb_{P_{W|\Ztrain,u}}[L_t(W|\Ztest)] \label{eq:modified_loss}
\end{align} is the average per-task training loss over the base-learner.

We emphasize that the meta-training loss \eqref{eq:metatrainingloss_sep} can be computed by the meta-learner and used as a criterion to select the meta-learning procedure \eqref{eq:meta_learner} since it is obtained from the meta-training data $\mset$. We also note that the rationale of splitting training and test sets is that the average training loss $L^{\trte}_t(u|Z^m_i)$ is an unbiased estimate of the corresponding average generalization loss $\Ebb_{P_{W|\Ztrain_i,u}}[L_g(W|T_i)]$.

The true goal of the meta-learner is to minimize the 
 \textit{meta-generalization loss},
\begin{align}
\Lm_g^{\trte}(u)=\Ebb_{P_{T,\Ztrain}} \Ebb_{P_{W|\Ztrain,u}} \bigl[L_g(W|T) \bigr]
 \label{eq:meta_testlloss},
\end{align} where $P_{T,\Ztrain}=P_T P_{\Ztrain|T}$ and $L_g(W|T)$ is as defined in \eqref{eq:genloss}. Unlike the meta-training loss \eqref{eq:metatrainingloss_sep}, the meta-generalization loss is evaluated on a new, meta-test task $T$ and on the corresponding training data $\Ztrain$. We distinguish the meta-generalization loss and meta-training loss by the subscripts $g$ and $t$, respectively in \eqref{eq:meta_testlloss} and \eqref{eq:metatrainingloss_sep}. The difference between the meta-generalization loss \eqref{eq:meta_testlloss} and the meta-training loss \eqref{eq:metatrainingloss_sep}, known as the \textit{meta-generalization gap}, is defined as
\begin{align}
\Delta \Lscr^{\trte} (u|\mset)=\Lm_{g}^{\trte}(u)- \Lscr_t^{\trte}(u|\mset).\label{eq:gap_1}
\end{align}
The quantity of interest to us is the average meta-generalization gap, defined as
\begin{align}
\Ebb_{P_{\mset,U}}\bigl[\DLm^{
\trte} (U|\mset)\bigr], \label{eq:metagen_gap2}
\end{align}where the expectation is with respect to the joint distribution $P_{\mset,U}= P_{\mset}P_{U|\mset}$, of the meta-training set $\mset$ and of the hyperparameter $U$. Note that $P_{\mset}$ is the marginal of the joint distribution $\prod_{i=1}^N P_{T=T_i}P_{Z^M|T=T_i}$.

Intuitively, if the meta-generalization gap is small, on average or with high probability, the meta learner can take the performance \eqref{eq:metatrainingloss_sep} on the meta-training data as a reliable measure of the accuracy of the inferred hyperparameter vector in terms of the meta-generalization loss \eqref{eq:meta_testlloss}. Furthermore, data-dependant bounds on the meta-generalization gap can be used as regularization terms to avoid meta-overfitting. Meta-overfitting occurs when the meta-trained hyperparameter yields a small meta-training loss but a large meta-test loss due to an excessive dependence on the meta-training set \citep{baxter2000model}.
\subsubsection{Joint Within-Task Training and Test Sets}
\begin{figure}[h!]
\centering
\includegraphics[scale=0.3,clip=true, trim = 0in  0in 0in 0in]{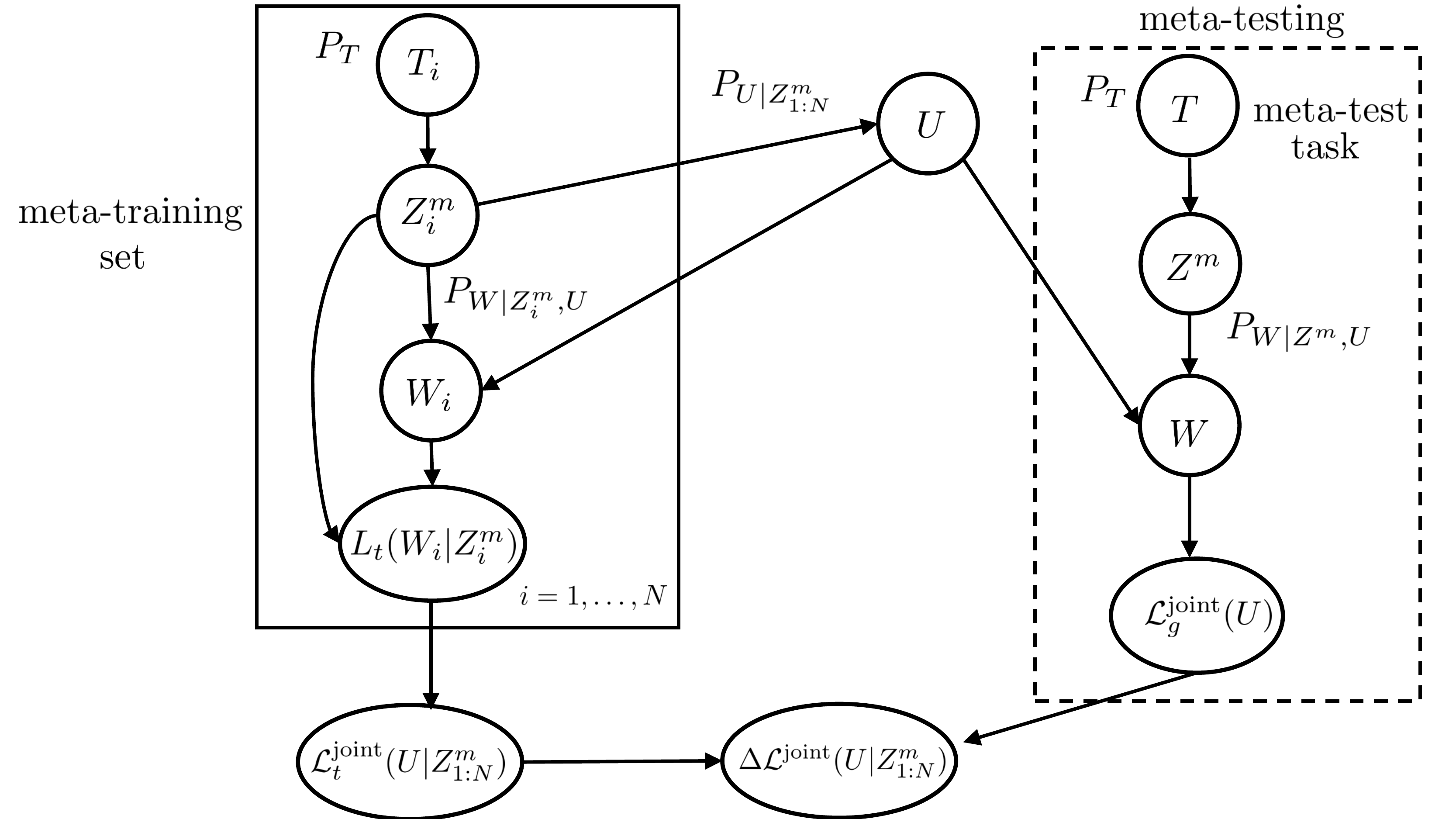} 
\caption{ Directed graph representing the variables involved in the definition of meta-generalization gap \eqref{eq:metagen_gap} for joint within-task training and testing sets. }\label{fig:BN_joint}
\end{figure}
In the second formulation of meta-learning, as illustrated in Figure~\ref{fig:BN_joint}, the entire data set $Z^m_i$ is used for within-task training and testing. Accordingly,
 the meta-learner computes the \textit{meta-training loss}
\begin{align}
\Lscr_t^{\jnt}(u|\mset)=\frac{1}{N}\sum_{i=1}^N L_t^{\jnt}(u|Z^m_i), \label{eq:meta_trainloss}
\end{align} where
\begin{align}
L_t^{\jnt}(u|Z^m)=\Ebb_{P_{W|Z^m,u}}[L_{t}(W|Z^m)]\label{eq:modifiedloss_joint}
\end{align} is the average per-task training loss. Note here that in evaluating the meta-training loss in \eqref{eq:meta_trainloss}, the data set $Z^m_i$ is used to infer model parameters $W$ and also to evaluate the per-task training loss. The expectation in \eqref{eq:modifiedloss_joint} is taken over the output of the base-learner $W$ given the hyperparameter vector $u$.  As discussed, the \textit{meta-generalization loss} for hyperparameter $u \in \Uscr$ is computed by randomly selecting a novel task $T \sim P_T$ as 
\begin{align}
\Lm_{g}^{\jnt}(u)=\Ebb_{P_{T,Z^m}} \Ebb_{P_{W|Z^m,u}}\bigl[ L_{g}(W|T)
\bigr],
\end{align} where $P_{T,Z^m}=P_T P_{Z^m|T}$ and $L_g(W|T)$ is as defined in \eqref{eq:genloss}.
 In a manner similar to \eqref{eq:gap_1}, the \textit{meta-generalization gap} for a task distribution $P_T$, data distribution $P_{Z^m|T}$, meta-learning algorithm $P_{U|\mset}$, and base-learner $P_{W|Z^m,U}$ is defined as
\begin{align}
& \DLm^{\jnt}(u|\mset)=\Lm_{g}^{\jnt}(u)-\Lscr_t^{\jnt}(u|\mset). \label{eq:metagen_gap}
\end{align}The average meta-generalization gap is then given as $\Ebb_{P_{\mset,U}}[\DLm^{\jnt}(U|\mset)]$ where the expectation is taken over all meta-training sets and over the output of the meta-learner.
 \section{Information-Theoretic Generalization Bounds for Single-Task Learning}\label{sec:singletaskbounds}
In this section, we review two information-theoretic bounds on the generalization gap \eqref{eq:gap_singletask} for conventional learning derived in \citep{xu2017information} and \citep{bu2019tightening}. The material covered in this section provides the necessary background for the analysis of the meta-generalization gap to be studied in the rest of the paper. Throughout this section, we fix a task $\tau \in \mathcal{T}$.  Since the generalization and meta-generalization gaps measure the deviation of empirical-mean random variables representing training and meta-training losses from reference values, we will make use of tools and definitions from large-deviation theory (see, e.g, \citep{wainwright2019high}). We discuss the key essential definitions below.
\subsection{Preliminaries}
 To start,
the cumulant generating function (CGF) of a random variable $X \sim P_X \in \Pscr(\Xscr)$ is defined as $\Lambda_X(\lambda)=\log \Ebb_{P_X}[e^{\lambda(X-\Ebb_{P_X}[X])}]$. If it is well-defined, the CGF $\Lambda_X(\lambda)$ is convex and it satisfies the equalities $\Lambda_X(0)=\Lambda'_X(0)=0$.
A random variable $X$ with finite mean, \ie, with $\Ebb_{P_X}[X] < \infty$, is said to $\sigma^2$-sub-Gaussian if its CGF is bounded as
\begin{align}
\Lambda_X(\lambda) \leq \frac{\lambda^2 \sigma^2}{2}, \quad \mbox{for all} \hspace{0.2cm} \lambda \in \Real. \label{eq:subGaussianity}
\end{align}
As a special case, if $X$ is bounded in the interval $[a,b]$, \ie, if the inequality $0 <a \leq X \leq b <\infty$ holds for some constants $a$ and $b$, then $X$ is $(b-a)^2/4$-sub-Gaussian.
\subsection{Mutual Information (MI) Bound}
We first present the Mutual Information (MI)-based upper bound obtained in \citep{xu2017information}.
   Key to this result is the following Assumption.
\begin{Assumption}\label{assum:1}
The loss function $l(w,Z)$ is $\delta_{\tau}^2$-sub-Gaussian under $Z \sim P_{Z|\tau}$ for all model parameters $w \in \Wscr$.
\end{Assumption}
In particular, if the loss function is bounded, \ie, if the inequalities $-\infty<a\leq l(w,z)\leq b <\infty$ hold for all   for  $w \in \Wscr$ and $z \in \Zscr$, Assumption~\ref{assum:1} is satisfied with $\delta_{\tau}^2=(b-a)^2/4$.
%
The main result is as follows.
\begin{Lemma}[\hspace{-0.03cm}\citep{xu2017information}]\label{cor:MI_singletask_subGaussian}
Under Assumption~\ref{assum:1}, the following bound on the generalization gap  holds for any base-learner $W\sim P_{W|Z^m,u}$
\begin{align}
| \Ebb_{P_{Z^m,W|\tau,u}}[\Delta L(W|Z^m,\tau)] |\leq \sqrt{\frac{2 \sigma^2}{m}I(W;Z^m)} \label{eq:MI_single_subGaussian}.
\end{align}
\end{Lemma}
The proof of Lemma~\ref{cor:MI_singletask_subGaussian} is based on a decoupling estimate Lemma, which is reported for completeness in Lemma~\ref{lem:decoupling}. We also note that the result in Lemma~\ref{cor:MI_singletask_subGaussian} can be extended to account for loss function $l(w,Z)$ with bounded CGF \citep{russo2016controlling}.

The bound \eqref{eq:MI_single_subGaussian} on the generalization gap is in terms of the mutual information $I(W;Z^m)$, which quantifies the overall dependence between the base-learner output $W$ and the input training data set $Z^m$. The mutual information in \eqref{eq:MI_single_subGaussian} is hence a measure of the \textit{sensitivity} of the base-learner output to the data set. Using the terminology in \citep{xu2017information}, if $ I(W;Z^m) \leq \epsilon$, the base-learner $P_{W|Z^m,u}$ is said to be $(\epsilon,P_{Z|\tau})$-MI stable, in which case the bound in \eqref{eq:MI_single_subGaussian} evaluates to $\sqrt{2 \delta_{\tau}^2\epsilon/m}$. The relationship between generalization and stability of a training algorithm is well-established \citep{shalev2014understanding}, and the result \eqref{eq:MI_single_subGaussian} amounts to a formulation of this link in information-theoretic terms.

 The traditional notion of algorithmic stability measures how much the base-learner output changes with the replacement of an individual training sample \citep{bousquet2002stability},\citep{shalev2010learnability}. In the next section, we review the bound in \citep{bu2019tightening} that translates this per-sample stability concept within an information-theoretic framework.
\subsection{Individual Sample MI (ISMI) Bound}
The MI-based bound in Lemma~\ref{cor:MI_singletask_subGaussian} has the 
disadvantage of being vacuous, \ie, $I(W;Z^m)=\infty$, for deterministic base-learning algorithms $P_{W|Z^m,u}$ defined on continuous parameter space $\Wscr$.
 An \textit{individual sample MI} (ISMI)-based  bound that address this shortcoming was introduced in \citep{bu2019tightening}. The ISMI bound borrows the standard algorithmic stability notion of sensitivity of the base-learner output to the replacement of any individual training sample \citep{devroye1979distribution, rogers1978finite}. Accordingly,
  the resulting bound is in terms of the MI between the trained parameter $W$ and each data point $Z_i$ of the training data set $Z^m$.
The bound, summarized in Lemma~\ref{cor:subGaussian_ISMI}, applies under  the following assumption.
\begin{Assumption}\label{assum:bu}
The loss function $l(w,z)$ satisfies either of the following two conditions:
\begin{itemize}
\item[$(a)$] Assumption~\ref{assum:1}, or
\item[$(b)$]  $l(W,Z)$ is a $\delta_{\tau}^2$-sub-Gaussian random variable when $(W,Z) \sim P_{W|u,\tau} P_{Z|\tau}$, where $P_{W|u,\tau}\in \Pscr(\Wscr)$ is the marginal of the joint distribution $P_{W|Z^m,u}P_{Z^m|\tau}$.
\end{itemize}
\end{Assumption}

We note that, in general, Assumption~\ref{assum:1} does not imply Assumption~\ref{assum:bu}$(b)$ (see \citep[Appendix C]{negrea2019information}), and vice versa (see \citep{bu2019tightening}). There are, however, loss functions $l(w,z)$ and relevant distributions for which both the assumptions hold, including the case of loss functions  $l(\cdot,\cdot)$ which  takes values in a bounded interval $[a,b]$.
\begin{Lemma}[\hspace{-0.02cm}\citep{bu2019tightening}]\label{cor:subGaussian_ISMI}
Under Assumption~\ref{assum:bu}, the following bound on the average generalization gap holds for any base-learner $P_{W|Z^m,u}$
\begin{align}
| \Ebb_{P_{Z^m,W|\tau,u}}[\Delta L(W|Z^m,\tau)]| \leq \frac{1}{m}\sum_{i=1}^m \sqrt{2 \sigma^2 I(W;Z_i)}  . \label{eq:ISMI_subGaussian}
\end{align}
\end{Lemma}

For a loss function satisfying Assumption~\ref{assum:1}, the ISMI bound \eqref{eq:ISMI_subGaussian} is tighter than
\eqref{eq:MI_single_subGaussian}, \ie,
\begin{align}
\frac{1}{m}\sum_{i=1}^m \sqrt{2\delta_{\tau}^2 I(W;Z_i)} \leq  \sqrt{\frac{2\delta_{\tau}^2}{m} I(W;Z^m)}. \label{eq:cor}
\end{align}
The inequality in \eqref{eq:cor} follows from the chain rule of mutual information and Jensen's inequality \citep{bu2019tightening}. 

\section{Information-Theoretic Generalization Bounds for Meta-Learning}
In this section, we first derive novel MI-based bounds on the meta-generalization gap with separate within-task training and test sets, as introduced in Section~\ref{sec:separatebounds}, and then we consider joint within-task training and test sets, as described in Section~\ref{sec:jointbounds}.
\subsection{Bounds on Meta-Generalization Gap with Separate Within-Task Training and Test Sets}\label{sec:separatebounds}
In this section, we present two novel MI-based bounds on the meta-generalization gap \eqref{eq:metagen_gap2} for the setup with separate within-task training and testing sets. The first is an MI-based bound, which is akin to Lemma~\ref{cor:MI_singletask_subGaussian}, and the second is an Individual Task MI (ITMI) bound, which resembles Lemma~\ref{cor:subGaussian_ISMI} for conventional learning.
\subsubsection{MI-Based Bound}
In order to derive the MI-based bound, we make the following assumption on $L^{\trte}_t(u|Z^m)$ in \eqref{eq:modified_loss}. Throughout, we use $P_{Z^m}$ to denote the marginal of the joint distribution $P_{T,Z^m}=P_TP_{Z^m|T}$.
\begin{Assumption}\label{assum:2}
For all $u \in \Uscr$, the average per-task training loss $L^{\trte}_t(u|Z^m)$ is $\sigma^2$-sub-Gaussian under $Z^m \sim P_{Z^m}$.
 \end{Assumption}

Distinct from the assumptions in Section~\ref{sec:singletaskbounds} on loss function $l(w,z)$, we note that Assumption~\ref{assum:2} is on the average per-task training loss $L_t^{\trte}(u|Z^m)$. This is because the loss function $l(w,z)$ satisfying Assumption~\ref{assum:1} do not in general guarantee sub-Gaussianity of $L_t^{\trte}(u|Z^m)$ with respect to $Z^m \sim P_{Z^m}$. However, if the loss function is bounded, Assumption~\ref{assum:2} can be easily verified to hold as given in the following lemma.
\begin{Lemma}\label{lem:sufficientcondition}
If the loss function $l(\cdot,\cdot)$ is $[a,b]-$bounded,
  then $L^{\trte}_t(\cdot|Z^m)$ is also $[a,b]$ bounded for all $Z^m \in \Zscr^m$. Consequently, $L^{\trte}_t(u|Z^m)$ is  $ (b-a)^2/4$-sub-Gaussian under $Z^m \sim P_{Z^m}$ for all $u \in \Uscr$.
\end{Lemma}

Under Assumption~\ref{assum:2}, the following theorem presents an upper bound on the meta-generalization gap \eqref{eq:metagen_gap2}.
\begin{Theorem}\label{thm:FMMIbound_different}
Let Assumption~\ref{assum:2} hold for the base-learner $P_{W|\Ztrain,u}$.
Then, for any meta learner $P_{U|\mset}$ such that the inequality $I(U;\mset)<\infty$ holds, we have the following bound on the average meta-generalization gap
\begin{align}
\biggl|\Ebb_{P_{\mset,U}}\bigl[\DLm^{
\trte} (U|\mset)\bigr]\biggr|&\leq \sqrt{\frac{2 \sigma^2}{N} I(U;\mset)}.
  \label{eq:FMMI_separate_subGaussian}
\end{align}
\end{Theorem}
\begin{proof}
See Appendix~\ref{app:3}.
\end{proof}


In order to prove Theorem~\ref{thm:FMMIbound_different}, one needs to overcome an additional challenge as compared to the derivation of bounds for learning reviewed in Section~\ref{sec:singletaskbounds}. In fact, the meta-generalization gap is caused by two distinct  sources of uncertainty: $(a)$ \textit{environment-level uncertainty} due to finite number $N$ of observed tasks, and $(b)$ \textit{within-task uncertainty} resulting from the finite number $m$ of per-task data samples. Our proof approach involves applying the single-task MI-based bound in Lemma~\ref{cor:MI_singletask_subGaussian} to bound the effect of both sources of uncertainties.

 Towards this, we start by introducing the average training loss for the randomly selected meta-test task as \begin{align}
\Lm^{\trte}_{g,t}(u)&=\Ebb_{P_{T,Z^m}}[L^{\trte}_t(u|Z^m)]. \label{eq:auxilliary_test}
\end{align}The subscript $g,t$ denote that the loss is generalization ($g$) with expectation over $P_{T,Z^m}$ at the environment level, and training ($t$) at the task level with $L^{\trte}_t(u|Z^m)$. Note that this differs from the meta-test loss $\Lm_{g}^{\trte}(u)$ in \eqref{eq:meta_testlloss} in that the per-task loss is evaluated in \eqref{eq:auxilliary_test} on the training set. With this definition the meta-generalization gap can be decomposed as
\begin{align}
&\Ebb_{P_{\mset,U}}\bigl[\DLm^{
\trte} (U|\mset)\bigr] \non \\&=\Ebb_{P_{\mset,U}} \biggl[(\Lm_{g}^{\trte}(U)-\Lm^{\trte}_{g,t}(U))+(\Lm^{\trte}_{g,t}(U)-\Lm^{\trte}_t(U|\mset))  \biggr]. \label{eq:decomposition}
\end{align} 
In \eqref{eq:decomposition}, the second difference $\Lm^{\trte}_{g,t}(U)-\Lm^{\trte}_t(U|\mset)$, corresponds to the environment-level uncertainty and arises from the observation of a finite number $N$ of tasks. In fact, as $N$ increases, the meta-training loss $\Lm^{\trte}_t(u|\mset)$ almost surely tends to $\Lm^{\trte}_{g,t}(u)$ by the law of large numbers. However, the average $\Ebb_{P_{\mset,U}}\bigl[ \Lm^{\trte}_{g,t}(U)-\Lm^{\trte}_t(U|\mset)\bigr]$ is not equal to zero in general for finite values of $N$. The within-task generalization gap is instead measured by the difference $\Lm_{g}^{\trte}(u)-\Lm^{\trte}_{g,t}(u)$. In the setup under study with separate within-task training and test sets, this term equals zero since as we discussed, the average empirical loss $L^{\trte}_t(u|Z^m_i)$ is an unbiased estimate of the corresponding average test loss $\Ebb_{P_{W|\Ztrain_i,u}}[L_g(W|T_i)]$ (cf. \eqref{eq:auxilliary_test} ). This is no longer true for joint within-task training and test sets, as we discuss in Section~\ref{sec:jointbounds}.
 
The decomposition approach adopted here follows the main steps of the bounding techniques introduced in \citep[equation (6)]{maurer2005algorithmic}.
In contrast, the PAC-Bayesian bounds in \citep{amit2018meta, rothfuss2020pacoh} rely on a different decomposition of the meta-generalization gap. The environment and within-task generalization gaps are then separately bounded in high probability, and are combined via  union bound to obtain the required PAC-Bayesian bounds.

The bound \eqref{eq:FMMI_separate_subGaussian} relates the meta-generalization gap to the information-theoretic stability of the meta-training procedure. As first introduced here, this stability is measured by the MI $I(U;\mset)$ between the hyperparameter $U$ and the meta-training data set $\mset$,
in a manner similar to the MI-based bounds in Lemma~\ref{cor:MI_singletask_subGaussian} for conventional learning. Importantly, as we will discuss in Section~\ref{sec:jointbounds}, this direct parallel between learning and meta-learning no longer applies with joint within-task training and test data sets.

\subsubsection{ITMI Bound}\label{sec:ITMI_separate}
We now present the ITMI bound, which holds under the following assumption.
\begin{Assumption}\label{assum:bu_metaseparatesupport}
Either of the following assumptions on the average per-task training loss $L^{\trte}_t(u|Z^m)$ holds:
\begin{itemize}
\item[$(a)$] $L^{\trte}_t(u|Z^m)$ satisfies Assumption~\ref{assum:2}, or
\item[$(b)$] $L^{\trte}_t(U|Z^m)$ is $\sigma^2$-sub-Gaussian under $(U,Z^m) \sim P_U P_{Z^m}$, where $P_{U}$ is the marginal of the joint distribution $P_{\mset,U}$ and $P_{Z^m}$ is the marginal of the joint distribution $P_{T,Z^m}$.
\end{itemize}
 \end{Assumption}
 
Assumption~\ref{assum:bu_metaseparatesupport} can be seen to be implied by the sufficient conditions in Lemma~\ref{lem:sufficientcondition}.
\begin{Theorem}\label{thm:ITMI_separate}
Let Assumption~\ref{assum:bu_metaseparatesupport} hold for the base-learner $P_{W|\Ztrain,U}$.
Then, for any meta learner $P_{U|\mset}$, the following bound on the meta-generalization gap \eqref{eq:metagen_gap2} holds
\begin{align}
\biggl|\Ebb_{P_{\mset,U}}\bigl[\DLm^{
\trte} (U|\mset)\bigr] \biggr|&\leq \frac{1}{N}\sum_{i=1}^N \sqrt{2 \sigma^2 I(U;Z_i^m)}. \label{eq:separate_ITMI_subGaussian}
\end{align}
where the MI $I(U;Z^m_i)$ is computed with respect to the joint distribution $P_{Z^m_i,U}$ obtained by marginalizing the probability distribution $P_{\mset,U}$.
\end{Theorem}
\begin{proof}
See Appendix~\ref{app:3}.
\end{proof}

As can be seen from \eqref{eq:separate_ITMI_subGaussian}, the ITMI bound on the meta-generalization gap is in terms of the MI $I(U;Z_i^m)$ between the output $U$ of the meta learner and each per-task data set $Z_i^m$. This, in turn, quantifies the sensitivity of the meta learner output to the replacement of a single per-task data set.  Moreover, under Assumption~\ref{assum:2}, the ITMI bound \eqref{eq:separate_ITMI_subGaussian} yields a tighter bound than the MI-based bound \eqref{eq:FMMI_separate_subGaussian}. This can be seen from the following sequence of relations
\begin{subequations}
\begin{equation}
\sqrt{\frac{1}{N}I(U;\mset)}=\sqrt{\frac{1}{N}\sum_{i=1}^N I(U;Z^m_i|Z^m_{(i-1)})}
\end{equation}
\begin{equation} \qquad\qquad \stackrel{(a)}{\geq} \sqrt{\frac{1}{N}\sum_{i=1}^N I(U;Z^m_i)} 
\end{equation}
\begin{equation}
\qquad \qquad \stackrel{(b)}{\geq}  \frac{1}{N}\sum_{i=1}^N \sqrt{ I(U;Z^m_i)}, \label{eq:ITMI-MIcomparison}
\end{equation}
\end{subequations}
where $Z^m_{(i-1)}=(Z^m_1,\hdots, Z^m_{i-1})$; $(a)$ follows since $Z^m_i$ is independent of $Z^m_{(i-1)}$; and $(b)$ follows from Jensen's inequality.
\subsection{Bounds on Generalization Gap with Joint Within-Task Training and Test Sets}\label{sec:jointbounds}
We now derive MI and ITMI-based bounds on the meta-generalization gap in \eqref{eq:metagen_gap} for the case with joint within-task training and test sets. As we will see, the key difference with respect to the case with separate within-task training and test sets is that the uncertainty due to finite number of per-task samples, measured by the second term in the decomposition \eqref{eq:decomposition}, contributes in a non-negligible way to the meta-generalization gap. Since there is no split into separate  within-task training and test sets, the average training loss with respect to the learning algorithm is given by $L_t^{\jnt}(u|Z^m)$ in \eqref{eq:modifiedloss_joint}. 
\subsubsection{MI-based Bound}
In order to derive the MI-based bound, we make the following assumptions.
\begin{Assumption}\label{assum:FMMIjoint}
We consider the following assumptions.
\begin{itemize}
\item[(a)] For each task $\tau \in \mathcal{T}$, the loss function $l(w,Z)$ satisifies Assumption~\ref{assum:1}, and
\item[(b)]The average per-task training loss $L_t^{\jnt}(u|Z^m)$ in \eqref{eq:modifiedloss_joint} is $\sigma^2$-sub-Gaussian for all $u \in \Uscr$ when $Z^m \sim P_{Z^m}$. 
\end{itemize}
\end{Assumption}
An easily verifiable sufficient condition for the above assumption to hold is the boundedness of loss function $l(w,z)$, which follows in a manner similar to Lemma~\ref{lem:sufficientcondition}.
\begin{Theorem}\label{thm:FMMIbound_same}
Let Assumption~\ref{assum:FMMIjoint} hold for a base-learner $W \sim P_{W|Z^m,U}$. 
  Then, 
for any meta learner $P_{U|\mset}$, we have the following bound on the meta-generalization gap \eqref{eq:metagen_gap}
\begin{align}
\biggl| \Ebb_{P_{\mset,U}}[\DLm^{\jnt}(U|\mset)]\biggr|&\leq
\sqrt{\frac{2 \sigma^2}{N}I(U;\mset)}+\Ebb_{P_T}\biggl[ \sqrt{\frac{2 \delta_T^2}{m} I(W;Z^m|T=\tau)}\biggr]. \label{eq:MI_joint_subGaussian}
\end{align}
where
 the MI $I(W;Z^m|T=\tau)$ is evaluated with respect to the distribution $P_{Z^m,W|T=\tau}$ obtained by marginalizing the joint distribution $P_{W|Z^m,U} P_{\mset,U} P_{Z^m|T=\tau}$.
\end{Theorem}
\begin{proof}
See Appendix~\ref{app:1}.
\end{proof}

With joint within-task training and test sets, the bound \eqref{eq:MI_joint_subGaussian} on the meta-generalization gap contain the contributions of  two mutual informations. The first, $I(U;\mset)$, quantifies the sensitivity of the meta learner output $U$ to the meta-training data set $\mset$. This term also appeared in the bound  \eqref{eq:FMMI_separate_subGaussian} with separate within-task training and test sets. Decomposing the meta-generalization gap in a manner analogous to \eqref{eq:decomposition}, it corresponds to a bound on the average of the second difference. The second contribution, $I(W;Z^m|T=\tau)$, quantifies the sensitivity of the output of the base-learner $P_{W|Z^m,U}$ to the data set $Z^m$ of the meta-test task $T$, when the hyperparameter is randomly selected by the meta-learner $P_{U|\mset}$ using the meta-training set $\mset$. This second term is in line with the single-task generalization gap bounds \eqref{eq:MI_single_subGaussian}, and it bounds the corresponding first difference in the decomposition \eqref{eq:decomposition}.

We finally note that the dependence of the bound in \eqref{eq:MI_joint_subGaussian} on the number of tasks $N$ and per-task samples $m$ is of  the order $1/\sqrt{N}+1/\sqrt{m}$. Meta-generalization bounds with similar dependence have been derived in \citep{amit2018meta} using PAC-Bayesian arguments. The bounds on excess risk for representation learning also follow similar order of dependence on $N$ and $m$ (c.f \citep[Thm.~2]{maurer2016benefit}).

\subsubsection{ITMI Bound on \eqref{eq:metagen_gap}}
For deriving the ITMI bound on the meta-generalization gap \eqref{eq:metagen_gap}, we assume the following.
\begin{Assumption}\label{assum:ITMIjoint}
Either of the following assumptions hold:
\begin{itemize}
\item[$(a)$] Assumption~\ref{assum:FMMIjoint} holds, or
\item[$(b)$] For each task $\tau \in \mathcal{T}$, the loss function $l(W,Z)$ is $\delta_{\tau}^2$-sub-Gaussian when $(W,Z)\sim P_{W|\tau}P_{Z|\tau}$, where
$P_{W|\tau}$ is the marginal of the joint distribution $P_{W|Z^m,U}P_{\mset,U} P_{Z^m|\tau}$.
The average per-task training loss $L_{t}^{\jnt}(U|Z^m)$ is  $\sigma^2$-sub-Gaussian when $(U,Z^m) \sim P_U P_{Z^m}$.\end{itemize}
\end{Assumption}
As in Section~\ref{sec:ITMI_separate},
Assumption~\ref{assum:ITMIjoint} can be seen to be implied by the sufficient conditions in Lemma~\ref{lem:sufficientcondition}.
\begin{Theorem}\label{thm:ITMIbound_same}
Under Assumption~\ref{assum:ITMIjoint}, for 
 any meta learner $P_{U|\mset}$, the following bound holds on the average meta-generalization gap
\begin{align}
 &\biggl| \Ebb_{P_{\mset,U}}[\DLm^{\jnt}(U|\mset)]\biggr|
 \leq  \frac{1}{N}\sum_{i=1}^N  
\sqrt{2 \sigma^2 I(U;Z^m_i)} +\Ebb_{P_T}\biggl[ \frac{1}{m}\sum_{j=1}^m\sqrt{2 \delta_{T}^2 I(W;Z_j|T=\tau)}\biggr], \label{eq:ITMI_joint_subGaussian}
\end{align}
where 
 the MI $I(U;Z^m_i)$ is evaluated with respect to $P_{Z^m_i,U}$ obtained by marginalizing $P_{\mset,U}$, and the MI $I(W;Z_j|T=\tau)$ is with respect to $P_{Z_j,W|T=\tau}$ obtained by marginalizing $P_{Z^m,W|T=\tau}$.
\end{Theorem}
\begin{proof}
See Appendix~\ref{app:1}.
\end{proof}
Similar to the bound in \eqref{eq:MI_joint_subGaussian}, the bounds on meta-generalization gap in \eqref{eq:ITMI_joint_subGaussian} are in terms of two types of mutual information, the first describing the sensitivity of the meta-learner and the second the sensitivity of the base-learner. Specifically, the MI $I(U;Z^m_i)$ quantifies the sensitivity of the output of the meta learner to per-task data set $Z^m_i$, and the MI $I(W;Z_j|T=\tau)$ measures the sensitivity of the output of the base-learner, $P_{W|Z^m,U}$ to each data sample $Z_i$ within the training set $Z^m$ of the meta-test task $T$. Moreover, it can be shown, in a manner similar to \eqref{eq:ITMI-MIcomparison}, that, under Assumption~\ref{assum:FMMIjoint}, the ITMI bound in \eqref{eq:ITMI_joint_subGaussian} is tighter than the MI bound in \eqref{eq:MI_joint_subGaussian}.
\subsection{Discussion on Bounds}
The bounds on the average meta-generalization gap obtained in this section generalizes the bounds for conventional single-task learning in Section~\ref{sec:singletaskbounds}. To see this, consider the task distribution $P_T=\delta(T-\tau)$ to be centered at  some task $ \tau \in \mathcal{T}$. Recall that in conventional learning, the hyperparameter $u$ is fixed a priori. As such, the mutual information $I(U;\mset)$ (for MI-based bounds) and $I(U;Z^m_i)$ (for ITMI-based bounds) vanishes. For the separate within-task training and test sets, this implies that that the average generalization gap is zero, which follows since the per-task test loss $L_t(W|\Ztest_i)$ is an unbiased estimate of per-task generalization loss $L_g(W|T_i)$. The MI- and ITMI-based bounds for the joint within-task training and test sets then reduce to
\begin{align}
\biggl| \Ebb_{P_{Z^m,W|\tau,u}}[\Delta L(W|Z^m,\tau)]  \biggr | \leq \sqrt{\frac{2 \delta_{\tau}^2}{m} I(W;Z^m)},
\end{align}
and 
\begin{align}
\biggl| \Ebb_{P_{Z^m,W|\tau,u}}[\Delta L(W|Z^m,\tau)]  \biggr | \leq \frac{1}{m}\sum_{j=1}^m\sqrt{2 \delta_{\tau}^2 I(W;Z_j)}
\end{align} respectively, where $I(W;Z^m)$ is evaluated with respect to the joint distribution $P_{W,Z^m|\tau,u}$ and $I(W;Z_j)$ with respect to $P_{W,Z_j|\tau,u}$.

The MI and ITMI-based bounds derived in this section point that a smaller correlation between hyperparameters and meta-training set and thus small mutual information $I(U;\mset)$ improves the meta-generalization gap, although this seems deleterious to performance. To clarify this contradiction, we would like to emphasize that these bounds quantify the difference between meta-generalization loss and empirical training loss, which in turn depends on the sensitivity of the meta-learner and base-learner to their input meta-training set and per-task training set, respectively. The mutual information terms in our bounds capture these sensitivities. Consequently, our bounds suggest that a meta-learner that is highly correlated to the input meta-training set (\ie, when $I(U;Z^m_{1:N})$ is large) does not generalize well (\ie, yields large meta-generalization gap). This property aligns with previous information-theoretic analysis for generalization in conventional learning \citep{xu2017information}. 

To the best of our knowledge, the MI-and ITMI-based bounds studied here are the first bounds on the average meta-generalization gap. As discussed in the introduction, these bounds are  distinct from the high-probability PAC and PAC-Bayesian bounds on the meta-generalization gap studied previously on meta-learning. Consequently, the bounds studied in this work are not directly comparable with the existing high-probability bounds.

Finally, we note that similarity between tasks is crucial to meta-learning. If the per-task data distributions $P_{Z|T=\tau}$ in the task environment are `closer' to each other, a meta-learner can efficiently learn the shared characteristics of tasks, and can generalize well to new tasks from the task environment. In our setting, the statistical properties of the task environment $(P_T,\{P_{Z|T=\tau}\}_{\tau \in \mathcal{T}})$ dictate this similarity. Although our MI and ITMI-based bounds do not explicitly capture this, we note that the properties of task environment are implicitly accounted for by the mutual information terms $I(U;\mset)$ and $I(U;Z^m_i)$ where the meta-training data set $\mset$ is generated from the task environment, and also by the sub-Gaussianity considerations in Assumption~\ref{assum:2},\ref{assum:bu_metaseparatesupport}, \ref{assum:FMMIjoint}, \ref{assum:ITMIjoint}. From preliminary studies, we believe that information-theoretic bounds that explicitly capture the impact of task similarity requires a different performance metric than the average meta-generalization gap considered here, and is left to future work.
\section{Applications}
In this section, we consider two applications of the information-theoretic bounds proposed in Section~\ref{sec:separatebounds}. The first, simpler, example concerns a parameter estimation problem for which an optimized meta-learner can be obtained in closed form. In contrast, the second application covers a broad class of iterative meta-training schemes.
\subsection{Parameter Estimation}\label{sec:example}
To illustrate the bounds on the meta-generalization gap derived in Section~\ref{sec:separatebounds}, we first consider the problem of prediction for a Bernoulli process with a `soft' predictor that uses only a few samples from the process, as well as meta-training data. Towards this, we consider an arbitrary discrete finite set of tasks $\mathcal{T}=\{\tau_1,\hdots, \tau_M\}$. The data distribution $P_{Z|T=\tau_k}$ for each task $\tau_k \in \mathcal{T}$, $k \in \{1,\hdots,M\}$, is given as $\Ber(\mu_{\tau_k})$ with mean parameter $\mu_{\tau_k}$.  The task distribution $P_T$ is then defined over the finite set of mean parameters  $\{\mu_{\tau_1},\hdots, \mu_{\tau_M}\}$. The base-learner uses training data, distributed  i.i.d. from $\Ber(\mu_{\tau_k})$ to determine the parameter $W_k$, which is used as a predictor of new observation $Z \sim \Ber(\mu_{\tau_k})$ at test time. The loss function is defined as $l(w,z)=(w-z)^2$, measuring the quadratic error between prediction and realized test input $z$. Note that the optimal (Bayes) predictor, computable in the ideal case of known distribution $P_{Z|T=\tau_k}$, is given as $W=\mu_{\tau_k}$.
We now distinguish the two cases with separate and joint within-task training and test sets.
\subsubsection{Separate within-task training and test sets}
The base-learner $P_{W|\Ztrain_{k},u}$ for task $\tau_k \in \mathcal{T}$, deterministically selects the prediction
\begin{align}
W_k=\alpha \Ztr_k+(1-\alpha)u, \label{eq:example_phi}
\end{align} where 
$
\Ztr_k=\frac{1}{\mtr}\sum_{j=1}^{\mtr} \Ztrain_{k,j}
$
is an empirical average over the training set $\Ztrain_{k,j}$, $u$ is a hyperparameter defining a bias that can be meta-trained, and $\alpha \in [0,1)$ is a fixed scalar. Here, $\Ztrain_{k,j}$ denote the $j$th data sample in the training set $\Ztrain_k$ of task $\tau_k$.  The bias term in \eqref{eq:example_phi} may help approximate the ideal Bayes predictor in the presence of limited data $\Ztrain_k$.

The objective of the meta-learner is to infer the hyperparameter $u$. 
 For a given meta-training data set $\mset$, comprising of data sets from $N$ tasks sampled according to $P_T$, the meta-learner can compute the empirical meta-training loss as
\begin{align}
\Lscr^{\trte}_t(u|\mset)=\frac{1}{N}\sum_{k=1}^N \frac{1}{\mte} \sum_{j=1}^{\mte} ( W_k -\Ztest_{k,j})^2, \label{eq:example_metaempricalloss_separate}
\end{align} where $\Ztest_{k,j}$ denote the $j$th example in the test set of $Z^m_k$, the $k$th sub-data set of $\mset$.
The meta-learner $P_{U|\mset}$ then deterministically selects the minimizing hyperparameter $u$ of the meta-training empirical loss function in \eqref{eq:example_metaempricalloss_separate}. This optimization yields
\begin{align}
U=\frac{(1-\alpha)^{-1}}{N}\biggl(\sum_{k=1}^N \Zte_k-\alpha \Ztr_k\biggr), \label{eq:theta_separate}
\end{align}where $\Zte_k=\sum_{j=1}^{\mte} \Ztest_{k,j}/\mte$. Note that $\Zte_k$ and $\Ztr_k$ are binomial random variables and by \eqref{eq:theta_separate}, $U$ takes finitely many discrete values and is bounded as $-\alpha (1-\alpha)^{-1} \leq U \leq (1-\alpha)^{-1}$. 
The meta-test loss can be explicitly computed as 
\begin{align}
\Lscr_g^{\trte}(u)=(1-\alpha)^2\bigl(u^2-2u \Ebb_{P_T}[\mu_T] \bigr)+\Ebb_{P_T} \biggl[\alpha^2\bigl(\mu_T^2+\frac{\mu_T \bar{\mu}_T}{\mtr}\bigr)+\mu_T-2\alpha \mu_T^2 \biggr], \label{eq:example_metatestloss}
\end{align}
where $\bar{\mu}_T=1-\mu_T$, and the average meta-generalization gap evaluates to
\begin{align}
\Ebb_{P_{\mset,U}}[\Delta \Lscr^{\trte}(U|\mset)]
&=\frac{2(1+\alpha^2)}{N} {\rm Var}_T +\frac{2 \Ebb_{P_T}[\mu_T \bar{\mu}_T]}{N} \biggl( \frac{1}{\mte}+\frac{\alpha^2}{\mtr}\biggr)
 \label{eq:example_metagap_separate},
\end{align} where ${\rm Var}_T=\bigl(\Ebb_{P_T}[\mu_T^2]-(\Ebb_{P_T}[\mu_T])^2 \bigr)$ is the variance of $\mu_T$.

To compute the MI and ITMI-based bounds on the meta-generalization gap \eqref{eq:example_metagap_separate},
it is easy to verify that the average training loss $L_{t}^{\trte}(\cdot|Z^m)$ is bounded, \ie, $0\leq L_{t}^{\trte}(\cdot|Z^m) \leq (1+\alpha)^2$ for all $u \in \Uscr$ and $Z^m \in \Zscr^m$.
Thus,
Assumption~\ref{assum:2} for the MI bound and also Assumption~\ref{assum:bu_metaseparatesupport} for the ITMI bound hold with $\sigma^2=(1+\alpha)^4/4$.
For the MI bound, we note that since the meta-learner is deterministic, we have that $I(U;\mset)=H(U)$.
The ITMI bound \eqref{eq:separate_ITMI_subGaussian} is given as
\begin{align}
|\Ebb_{P_{\mset,U}}[\Delta \Lscr^{\trte}(U|\mset)]| \leq \frac{1}{N}\sum_{i=1}^N \sqrt{\frac{(1+\alpha)^4}{2}I(U;Z^m_i)}. \label{eq:exampleITMI_separate}
\end{align}
The information-theoretic measures in  \eqref{eq:exampleITMI_separate} can be evaluated numerically as discussed in Appendix~\ref{app:example}.
\begin{figure}[h!]
\centering
\includegraphics[scale=0.65,clip=true, trim = 2.7in  2in 2.8in 1.8in]{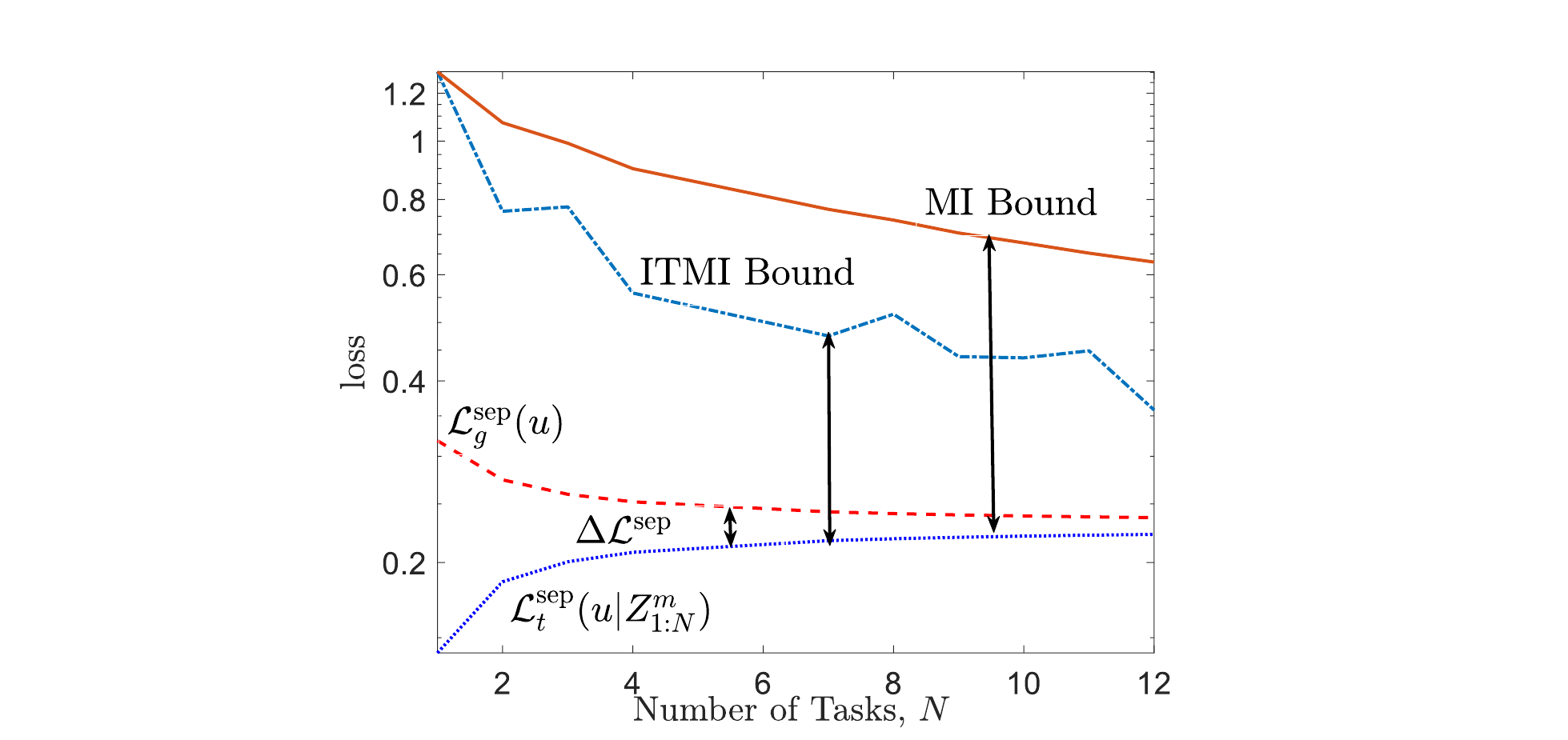} 
\caption{ Comparison of the MI bound in \eqref{eq:FMMI_separate_subGaussian} and ITMI based bound obtained in  \eqref{eq:exampleITMI_separate} with the meta-generalization gap for meta-learning with separate within-task training and test sets. The task environment is defined by $M=12$ tasks
    . Other parameters are set as $\alpha=0.15$, $\mtr=15$, $\mte=5$. }\label{fig:example_separate}
\end{figure}
For a numerical illustration,
Figure~\ref{fig:example_separate} plots the average of the meta-generalization loss \eqref{eq:example_metatestloss} and average meta-training loss \eqref{eq:example_avgtrain_sep} along with the ITMI bound in \eqref{eq:exampleITMI_separate} and MI bound in \eqref{eq:FMMI_separate_subGaussian}. It can be seen that the ITMI bound is tigher than MI bound and  correctly predicts the decrease in the meta-generalization gap as the number $N$ of tasks increases.

%
\subsubsection{Joint Within-Task Training and Testing sets}
\begin{figure}[h!]
\centering
\includegraphics[scale=0.7,clip=true, trim = 1.8in  1.5in 2in 1.8in]{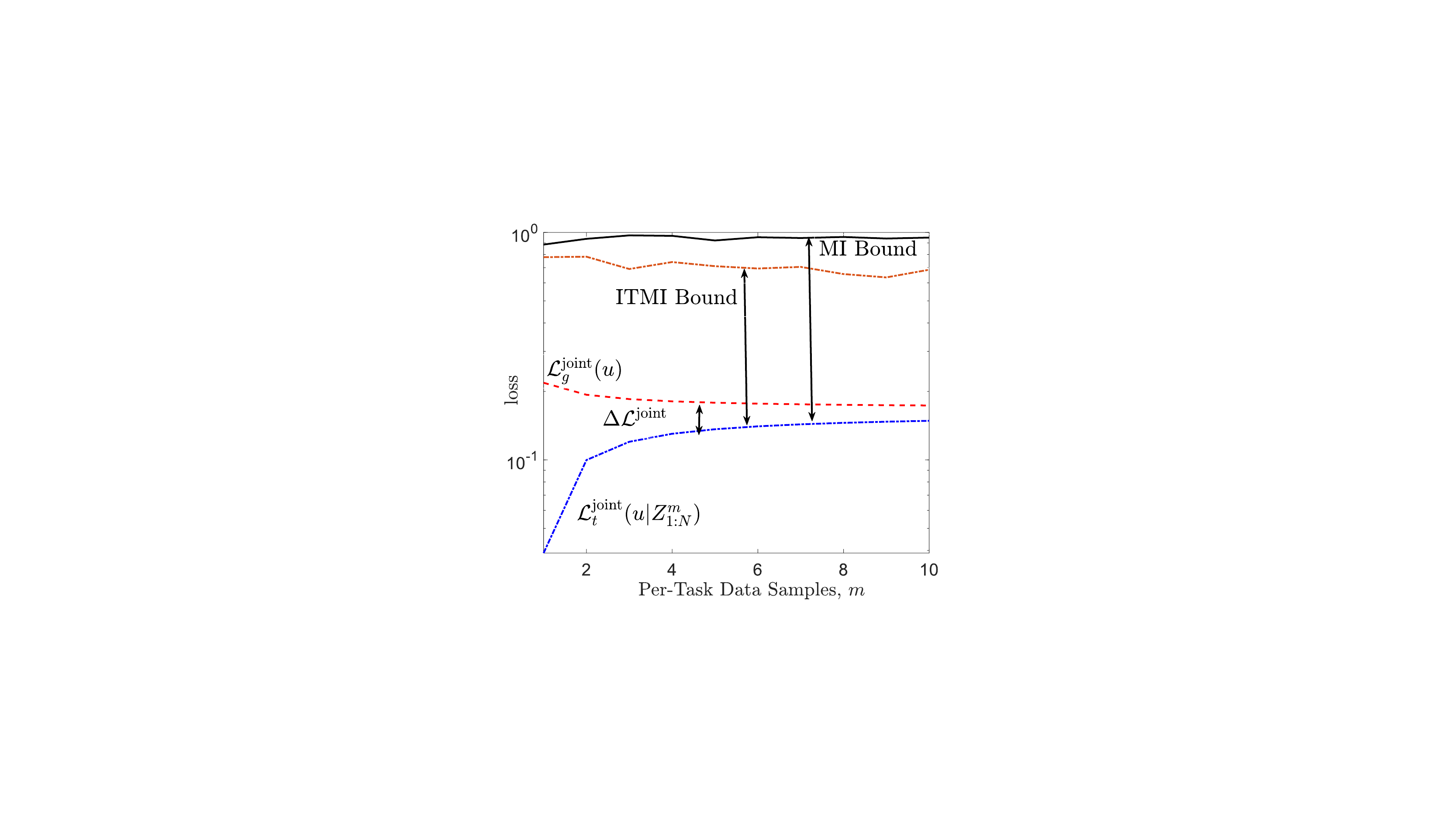} 
\caption{ Comparison of the MI and ITMI based bound obtained in \eqref{eq:exampleITMI_joint} with the meta-generalization gap for meta-learning with joint within-task training and test sets, as a function of the per-task data samples $m$ for $N=5$ and $\alpha=0.55$. The task environment is defined by $M=9$ tasks.
    }\label{fig:example_joint}
\end{figure}

We now consider the case with joint within-task training and test sets.
The base-learner $P_{W|Z^m_k,U}$ for task $\tau_k\in \mathcal{T}$ still uses the predictor \eqref{eq:example_phi}, but now the empirical average over the training set is given as
$
D_k=\sum_{j=1}^m Z^m_{k,j}/m.
$
As before,
the meta-learner $P_{U|\mset}$ deterministically selects the minimizing hyperparameter $u$ of the meta-training empirical loss function,
$L_{\mset}(u)=(1/N) \sum_{k=1}^N (1/m) \sum_{j=1}^{m} ( W_i -Z^m_{k,j})^2,$
 yielding
$
U=\frac{1}{N}\sum_{k=1}^N  D_k. $
As discussed in Appendix~\ref{app:example}, the meta-generalization loss for this example can also be explicitly computed and the meta-generalization gap bounds in \eqref{eq:MI_joint_subGaussian} and \eqref{eq:ITMI_joint_subGaussian} can be evaluated numerically. Figure~\ref{fig:example_joint} plots the average meta-generalization loss and average meta-training loss along with the MI bound in \eqref{eq:MI_joint_subGaussian} and ITMI bound in \eqref{eq:exampleITMI_joint}, as a function of per-task data samples $m$. The ITMI bound is seen to better reflect the decrease of the meta-training loss as a function of $m$.

\subsection{ Noisy Iterative Meta-Learning Algorithms}
Most meta-learning algorithms are built around a nested loop structure, with the inner loop applying the base-learner on the meta-training set and the outer loop updating the hyperparameters $U$. In this section, we focus on a vast class of such meta-learning algorithms in which the inner loop applies training procedures dependent on the current iterate of the hyperparameter, while the outer loop updates the hyperparameter using a stochastic rule. This class includes stochastic variants of state-of-the-art algorithms such as MAML \citep{finn2017model} and Reptile \citep{nichol2018first}. We apply the derived information-theoretic bounds to study the meta-generalization performance of the mentioned class of meta-training iterative stochastic rules by focusing on the case of separate within-task training and test sets here, which is assumed \eg, by MAML. The analysis for the setup with joint within-task training and test sets can also be carried out at the cost of a more cumbersome notation.

To start,
let $U^j \in \Real^d$ denote the hyperparameter vector at outer iteration $j$, with $U^0 \in \Real^d$ being an arbitrary initialization. For example, in MAML, the hyperparameter $U$ defines the initial iterate used by each base-learner in the inner loop to update the model parameter $W_{\tau}$ corresponding to task $\tau$. At each iteration $j \geq 1$, we randomly select a mini-batch of task indices $K_j \subseteq [1,\hdots,N]$ from the meta-training data $\mset$, obtaining the corresponding data set $Z^m_{K_j}= (\Ztrain_{K_j},\Ztest_{K_j}) \subseteq \mset$, where $\Ztrain_{K_j}=\{\Ztrain_k\}_{k \in K_j}$ and $\Ztest_{K_j}=\{\Ztest_k\}_{k\in K_j}$ are the separate training and test sets for the selected tasks. For each index $k\in K_j$, in the inner loop, the base-learner selects the model parameter $W_k^j$ as a, possibly stochastic, function
\begin{align}
W_k^j=g(U^{j-1},\Ztrain_{k}).
 \label{eq:baselearner_iterative}
\end{align}
For instance, in MAML, the function $g(U^{j-1},\Ztrain_k) \in \Real^{d}$ in \eqref{eq:baselearner_iterative} represents the output of an SGD procedure that starts from initialization $U^{j-1}$ and uses the task training data $\Ztrain_k$ to iteratively update the model parameters, producing the final iterate $W^j_k$.
We denote as $W_{K_j}=\{W^j_k\}_{k \in K_j}$ the collection of the base-learners' outputs for all task indices $k \in K_j$ at outer iteration $j$.

In the outer loop, the meta learner uses
the task-specific adapted parameters $W_{K_j}$ from the inner loop and the meta-test set $\Ztest_{K_j}$ to update the past iterate $U^{j-1}$ according to the general update rule
\begin{align}
U^j = F(U^{j-1})+\beta_j G(U^{j-1},W_{K_j},\Ztest_{K_j})+\xi_j,\label{eq:updaterule} 
\end{align}
where $F(\cdot)$ and $G(\cdot,\cdot,\cdot)$ are arbitrary deterministic functions; $\beta_j$ is the step-size; and $\xi_j \sim \mathbf{\mathcal{N}}(0, \gamma_j^2 I_d)$ is an isotropic Gaussian noise, independently drawn for $j=1,2,\hdots,$. As an example, in MAML, the function $F(\cdot)$ is the identity function and function $G(\cdot,\cdot,\cdot)$ equals the gradient of the empirical  loss $1/|K_j|\sum_{k \in K_j}L_{t}^{\trte}(W^j_{k}|\Ztest_k)$ in \eqref{eq:metatrainingloss_sep} with respect to $U^{j-1}$. Note, however, that MAML does not add noise, \ie, $\gamma_j^2=0$ for all $j$.

\begin{figure}[h!]
\centering
\includegraphics[scale=0.3,clip=true, trim = 0in  0in 0in 0in]{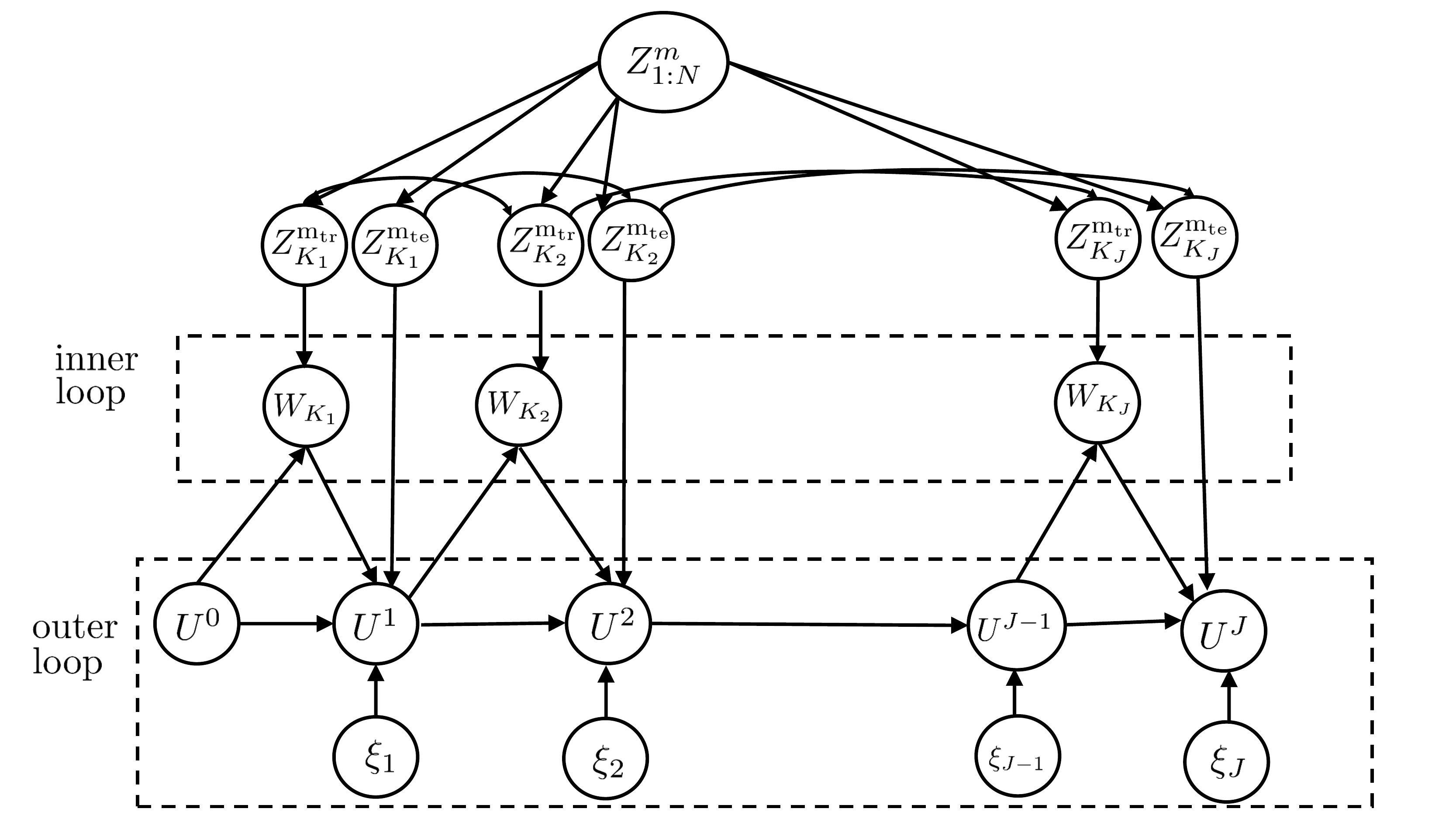} 
\caption{ A graphical model representation of the variables involved in the Definition of noisy iterative algorithms. }\label{fig:iterative_algorithm}
\end{figure}

The final output of the meta-learning algorithm is then defined as an arbitrary function
$U=f(U^1,\hdots,U^J),$ of all iterates. Examples of function $f$ include the last update $f(U^1,\hdots,U^J)=f(U^J)$ and average of the updates $f(U^1,\hdots,U^J)=1/J\sum_{j=1}^J U^j$. A graphical model representation of the variables involved is shown in Figure~\ref{fig:iterative_algorithm}.

 We now derive an upper bound on the meta-generalization gap for the general class of iterative meta-learning algorithm satisfying \eqref{eq:baselearner_iterative} - \eqref{eq:updaterule} under the following assumptions.
\begin{Assumption}\label{assum:iterative}
\begin{itemize}
\item[(1)]For the base-learner given in \eqref{eq:baselearner_iterative}, the average training loss  $L^{\trte}_{t}(u|Z^m)$ in \eqref{eq:modified_loss} is $\sigma^2$-sub-Gaussian for all $u \in \Uscr$ when $Z^m \sim P_{Z^m}$;
\item[(2)] The meta-training data set $Z^m_{K_j}$ sampled at each iteration $j$ is conditionally independent of the history of model-parameter vectors $\{W_{K_i}\}_{i=1}^{j-1}$ and hyperparameter $U^{(j-1)}=(U^1,U^2,\hdots,U^{j-1})$, \ie,
\begin{align}
P_{Z^m_{K_j} |\{Z^m_{K_i}\}_{i=1}^{j-1},\mset,U^{(j-1)},\{W_{K_i}\}_{i=1}^{j-1}}=P_{Z^m_{K_j} |\{Z^m_{K_i}\}_{i=1}^{j-1},\mset}; \label{eq:sampling}
\end{align}
\item[(3)] The meta-parameter update function $G(\cdot,\cdot,\cdot)$  is uniformly bounded, \ie, 
$||G(\cdot,\cdot,\cdot)||_2  \leq L$
for some $L>0$.
\end{itemize}
\end{Assumption}
\begin{Lemma}\label{lem:noisyalgorithm}
Under Assumption~\ref{assum:iterative}, the following upper bound on the meta-generalization gap \eqref{eq:metagen_gap2} holds for the class of noisy iterative meta-training algorithms \eqref{eq:baselearner_iterative}-\eqref{eq:updaterule} 
\begin{align}
\Ebb_{P_{\mset,U}}[\Delta \Lscr^{\trte}(U|\mset)] &\leq \sqrt{\frac{2\sigma^2}{N} \sum_{j=1}^J \frac{d}{2}\log \biggl(1+\frac{\beta_j^2L^2}{d\gamma_j^2  }\biggr)}. \label{eq:iterativebound}
\end{align}
\end{Lemma}
\begin{proof}
See Appendix~\ref{app:iterative}.
\end{proof}

The bound in \eqref{eq:iterativebound} has the same form as the generalization gap derived in \citep{pensia2018generalization} for conventional learning. From \eqref{eq:iterativebound}, the generalization gap can be reduced by increasing the variance $\gamma_j^2$ of the injected Gaussian noise. In particular, the meta-generalization gap depends on the ratios $\beta_j^2/\gamma_j^2$ between squared step size $\beta_j^2$ and variance $\gamma_j^2$. For example,
SGLD sets $\gamma_j=\sqrt{\beta_j}$, and a step size $\beta_j$ decaying over time according to the standard Robbins-Monro conditions in order to ensure convergence of the output samples to the generalized posterior distribution of the hyperparameters \citep{welling2011bayesian}.


\textit{Example}:
To illustrate bound \eqref{eq:iterativebound}, we now consider a simple logistic regression problem that generalizes the example studied in Section~\ref{sec:example}. Accordingly, each data point $Z$ corresponds to labelled data $Z=(X,Y)$, where $X \in \{0,1\}^d$ represents the input vector and $Y \in \{0,1\}$ represents the corresponding binary label. The data distribution $P_{Z|\tau_k}=P_{X|\tau_k} P_{Y|X,\tau_k}$ for each task $\tau_k \in \mathcal{T}=\{\tau_1,\hdots,\tau_M\}$ is such that $X \sim P_{X|\tau_k}$ is a $d$-dimensional Bernoulli vector obtained via $d$ independent draws from $\Ber(\nu)$ and  $Y$ is distributed as $Y \sim \Ber(\phi(\mu_{\tau_k}^TX))$, where  $\phi(a)=1/(1+\exp(-a))$ is the sigmoid function and $\mu_{\tau_k} \in \Real^d$, with   $||\mu_{\tau_k}||_2 \leq 1$. The task distribution $P_T$ then defines a distribution over the parameter vectors $\{\mu_{\tau_1},\hdots, \mu_{\tau_M}\}$. The base-learner uses training data generated i.i.d. from $P_{Z|\tau_k}$  to obtain a prediction $w$ of the parameter vector $\mu_{\tau_k}$ for task $\tau_k \in \mathcal{T}$.
The loss function is taken as the quadratic error   $l(w,z)=(\phi(w^Tx)-y)^2.$
 
At each iteration $j$, starting from initialization point $U^{j-1}$, the base-learner in \eqref{eq:baselearner_iterative} uses a one-step projected gradient descent algorithm on the training data set $\Ztrain_k$ to obtain the prediction $W^j_k$ as
\begin{align}
W^j_k={\rm proj}_{\Wscr}\biggl(U^{j-1}-\alpha \nabla_{w}L_{t}^{\trte}(w|\Ztrain_k) \big |_{w=U^{j-1}}  \biggr), \label{eq:base_learner_iterativeexample}
\end{align}where $\alpha>0$ is the step-size, $\Wscr=\{w \in \Real^d \bigl|\hspace{0.2cm} ||w||_2\leq 1\}$ is the set of feasible model parameters and ${\rm proj}_{\Ascr}(b)=\frac{1}{2}\min_{a \in \Ascr}||a-b||^2_2$ is the projection operator.
The meta-learner \eqref{eq:updaterule} updates the initialization vector according to the noisy gradient descent rule
\begin{align}
U^j=U^{j-1}-\beta_j \biggl( \frac{1}{|K_j|} \sum_{k=1}^{|K_j|}\nabla_{w}L_{t}^{\trte}(w|\Ztest_k) \big |_{w=W_k^{j} }\biggr) +\xi_j, \label{eq:example_iterative_outerloop}
\end{align}
where $\beta_j$ is the step-size; and $\xi_j \sim \Nscr(0,\gamma_j^2 I_d)$ is isotropic Gaussian noise. This update rule corresponds to performing an First Order MAML (FOMAML) \citep{finn2017model} with the addition of noise.

For this problem, it is easy to verify that  Assumption~\ref{assum:iterative} is satisfied, since
 the loss function $l(\cdot,\cdot)$ is bounded in the interval $[0,1]$, whereby $L^{\trte}_{t}(u|Z^m)$ is also $[a,b]$-bounded. We also have the inequality
\begin{align}
\left \lVert \frac{1}{|K_t|} \sum_{i=1}^{|K_t|}\nabla_{w}L_{\Ztest}(w) \big |_{w=W_i^{t} }\right \rVert_2 \leq 2\sqrt{d} e^{\sqrt{d}} \triangleq L.
\end{align}
The MI bound in \eqref{eq:iterativebound} then evaluates to
\begin{align}
\Ebb_{P_{\mset,U}}[\Delta \Lscr(U|\mset)] &\leq \sqrt{\frac{1}{2N} \sum_{j=1}^J \frac{d}{2}\log \biggl(1+\frac{4\beta_j^2 e^{2\sqrt{d}}}{\gamma_j^2  }\biggr)}. \label{eq:example_iterative_bound}
\end{align}
\begin{figure}[h!]
\centering
\includegraphics[scale=0.75,clip=true, trim = 2.5in  1.9in 0.3in 2in]{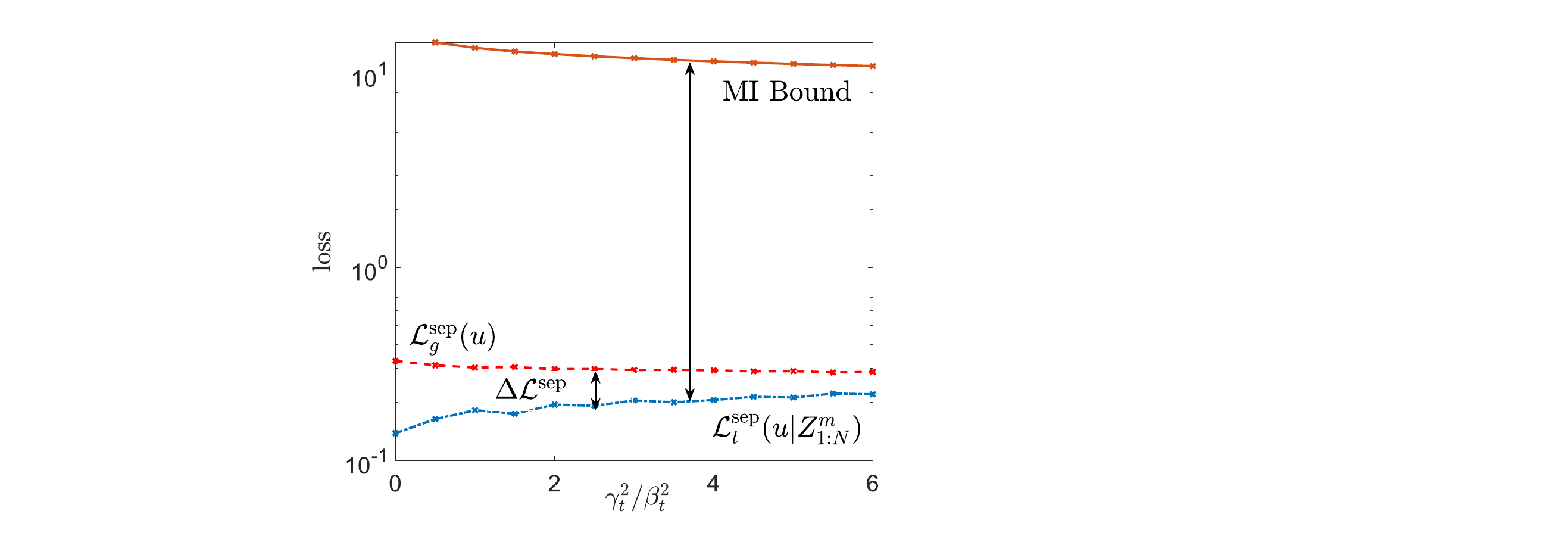}
            \caption{Comparison of the meta-generalization gap with the MI-based bound in \eqref{eq:example_iterative_bound} as function of the ratio $\gamma_t^2/\beta_t^2$. 
  }
          \label{fig:iterative_gapMI}
\end{figure}

We now evaluate the meta-training and meta-test loss, along with the bound \eqref{eq:example_iterative_bound} as a function of the ratio $\gamma_j^2/\beta_j^2$ in Figure~\ref{fig:iterative_gapMI}.
For the experiment, we considered a task environment of $M=20$ tasks with $\nu=0.4$, $d=3$, $N=4$ meta-training tasks with $\mtr=10$ training data samples and $\mte=5$ test data samples. For the inner-loop \eqref{eq:base_learner_iterativeexample}, we fixed step-size $\alpha=10^{-4}$ and for the outer-loop \eqref{eq:example_iterative_outerloop}, we set $|K_t|=N$, $\beta_j= 0.25$ and $T=200$ iterations. 
  
As suggested by Lemma~\ref{lem:noisyalgorithm}, the meta-generalization gap decreases with addition of noise. While the MI bound \eqref{eq:iterativebound} is generally loose, it correctly quantifies the dependence of the meta-generalization loss and the ratio $\gamma_j^2/\beta_j^2$, and it can hence serve as a useful meta-training criterion \citep{amit2018meta, yin2019meta}.
\section{Conclusions}
This work has presented novel information-theoretic upper bounds on the average generalization gap of meta-learning algorithms, thereby extending the well-studied information-theoretic approaches in conventional learning to meta-learning. The proposed bounds capture two sources of uncertainty -- environment-level uncertainty and within-task uncertainty -- and bound them via separate mutual information terms. Applications were also discussed with the aim of elucidating the use of the bounds to quantify meta-overfitting and guide the choice of the meta-inductive bias, \ie, the class of inductive biases. The derived bounds are amenable to further refinements such as those along the lines of \citep{asadi2018chaining,negrea2019information,steinke2020reasoning}. It would also be interesting to study the meta-generalization bounds on noisy iterative meta-learning algorithms using the tighter information-theoretic bounds such as \citep{bu2019tightening}, \citep{negrea2019information}. 
\funding{The authors have received funding from the European Research Council
(ERC) under the European Union’s Horizon 2020 Research and Innovation
Programme (Grant Agreement No. 725731). }

\appendixtitles{no} 
\appendix
\section{Decoupling Estimate Lemmas}
The proofs of the main results rely on the following decoupling estimate lemmas, which bound the difference in expectations under a change of measure from the joint $P_{X,Y}$ to the product of the marginals $P_X  P_Y$. In order to state the most general form of decoupling estimate lemmas, we first define a generalized sub-Gaussian random variable.
\begin{Definition}\label{def:generalizedsub-Gaussian}
A random variable $X$ is said to be $(\Psi_{+},\Psi_{-},b_{+},b_{-})$-\textit{generalized sub-Gaussian} if there exists convex functions $\Psi_{+}: \Real_{+}\rightarrow \Real$ and $\Psi_{-}:\Real_{+}\rightarrow \Real$ that satisfy the equalities $\Psi_{+}(0)=\Psi_{-}(0)=\Psi_{+}'(0)=\Psi_{-}'(0)=0$ and bound the CGF of $X$ as
\begin{subequations}
\begin{equation}
\Lambda_X(\lambda)\leq \Psi_{+}(\lambda), \qquad \mbox{for} \hspace{0.1cm} \lambda \in [0,b_{+})
\end{equation}
\begin{equation}
\Lambda_X(\lambda)\leq \Psi_{-}(-\lambda), \quad \mbox{for} \hspace{0.1cm} \lambda \in (b_{-},0],
\end{equation}
\end{subequations}
for some constants $0<b_{+}\leq \infty$ and $-\infty \leq b_{-}<0$.
\end{Definition}

For a $(\Psi_{+},\Psi_{-},b_{+},b_{-})$-generalized sub-Gaussian random variable, we also introduce the following standard definitions. First, 
the Legendre dual of  function $\Psi_{+}(\lambda)$ is defined as
\begin{align}
\Psi^{*}_{+}(x)=\sup_{\lambda \in [0,b_{+})}(\lambda x -\Psi_{+}(\lambda)). \label{eq:Legendredual}
\end{align}
It can be easily seen that $\Psi^{*}_{+}(\cdot)$ is a non-negative, convex, and non-decreasing function on $[0,\infty)$ with $\Psi^{*}_{+}(0)=0.$ Second, 
the inverse Legendre dual of function $\Psi_{+}(\lambda)$ is defined as $\Psi^{*-1}_{+}(y)=\inf\{x \geq 0: \Psi^{*}_{+}(x) \geq y\}$. This function is concave, and it can be equivalently written as \citep{bu2019tightening}
\begin{align}
\Psi^{*-1}_{+}(y)=\inf_{\lambda \in [0,b_{+})} \frac{y+\Psi_{+}(\lambda) }{\lambda}.\label{eq:inverseLegendre}
\end{align} Similar definitions and results apply for $\Psi_{-}(\cdot)$.

A $\sigma^2$-sub-Gaussian random variable $X$ is a generalized sub-Gaussian variable with $\Psi_{+}(\lambda)=\Psi_{-}(\lambda)=\lambda^2 \sigma^2/2$, $b_{+}=\infty$ and $b_{-}=-\infty$. Furthermore, the Legendre dual functions are given as $\Psi^{*}_{+}(x)=\Psi^{*}_{-}(x)=x^2/(2 \sigma^2)$, and the inverse Legendre dual functions  evaluate to 
\begin{align}
\Psi^{*-1}_{+}(y)=\Psi^{*-1}_{-}(y)=\sqrt{2\sigma^2 y}. \label{eq:subgaussian-general}
\end{align}

We are now ready to state the decoupling estimate lemmas.
\begin{Lemma}[Decoupling Estimate \citep{russo2016controlling}]\label{lem:decoupling}
Let $X \in \Xscr$ and $Y\in \Yscr$ be two jointly distributed random variables with joint distribution $P_{X,Y}$, and let $f(X,Y)$ be a real valued function such that $f(x,Y)$ is $(\Psi_{+},\Psi_{-},\infty,-\infty)$-generalized sub-Gaussian for all $x \in \Xscr$ when $Y \sim P_Y$. Then we have the following inequalities
\begin{align}
 \pm \biggl( \Ebb_{P_X  P_Y}[f(\Xt,\Yt)]-\Ebb_{P_{X,Y}}[f(X,Y)] \biggr)&\leq \Psi_{\mp}^{*-1}(I(X;Y)), \label{eq:decoupling}
\end{align}
where $(\Xt,\Yt)\sim P_X P_Y $.
\end{Lemma}

%

\begin{Lemma}[General Decoupling Estimate \citep{bu2019tightening}]\label{lem:general_decoupling}
Let $X \in \Xscr$ and $Y\in \Yscr$ be two jointly distributed random variables with joint distribution $P_{X,Y}$, and let 
 $f(X,Y)$ be a real valued function such that $f(X,Y)$ is a $ (\Psi_{+},\Psi_{-},b_{+},b_{-})$-generalized sub-Gaussian when $(X,Y)\sim P_X P_Y$. Then,
we have the inequality \eqref{eq:decoupling}.
\end{Lemma}

Note that in Lemma~A2 the random variables $X,Y$ are jointly distributed according to $P_{X,Y}$. Assuming that the function $f(X,Y)$ is generalized sub-Gaussian under $X \sim P_X$ and $Y \sim P_Y$ with $P_X$ and $P_Y$ being the marginals of $P_{X,Y}$,
 the lemma provides an upper bound on the  difference between average of $f(X,Y)$ when $(X,Y)$ is jointly distributed according to $P_{X,Y}$ and the average of $f(X,Y)$ when $(X,Y)$ is independent with $X \sim P_X$ and $Y \sim P_Y$. The resultant bound thus provides an estimate of the effect of decoupling of the joint distribution to its marginals with respect to function $f(X,Y)$.

\section{Proofs of Theorem~\ref{thm:FMMIbound_different} and Theorem~\ref{thm:ITMI_separate}}\label{app:3}
For the proof of Theorem~\ref{thm:FMMIbound_different}, we use the decomposition \eqref{eq:decomposition} of the meta-generalization gap into average environment-level and within-task generalization gaps as 
\begin{align}
&\Ebb_{P_{\mset,U}}[\Delta \Lscr^{\trte}(U|\mset)]\non \\ &=\Ebb_{P_{\mset,U}} \biggl[(\Lm^{\trte}_{g}(U)-\Lscr^{\trte}_{g,t}(U))+(\Lscr^{\trte}_{g,t}(U)-\Lscr_t^{\trte}(U|\mset))\biggr]\non \\
&=\Ebb_{P_T} \biggl[\Ebb_{P_{Z^m|T}P_{W,U,\mset|\Ztrain}}[\DL(W|\Ztest,T)]) \biggr]+\Ebb_{P_{\mset,U}}[\Lscr^{\trte}_{g,t}(U)-\Lscr_t^{\trte}(U|\mset)] \label{eq:auxiliary _metagap}
\end{align} 
where  \eqref{eq:auxiliary _metagap} follows since the average within-task generalization gap for a random meta-test task $\Ebb_{P_{\mset,U}} [\Lm^{\trte}_{g}(U)-\Lscr^{\trte}_{g,t}(U)]$ can be equivalently written as  $\Ebb_{P_T} \bigl[\Ebb_{P_{Z^m|T}P_{W,U,\mset|\Ztrain}}[\DL(W|T)]) \bigr]$, with  $\DL(W|\Ztest,T) =L_{P_{Z|T}}(W)-L_{t}(W|\Ztest)$ denoting the generalization gap of the meta-test task $T$, and the joint distribution
$P_{W,U,\mset|\Ztrain}$ factorizes as $P_{W|\Ztrain,U}P_{\mset,U}$.
To obtain an upper bound on the average meta-generalization gap $\Ebb_{P_{\mset,U}}[\Delta \Lscr^{\trte}(U|\mset)]$, we bound each of the two differences in \eqref{eq:auxiliary _metagap} separately.

We first bound the second difference in \eqref{eq:auxiliary _metagap}
$\Ebb_{P_{\mset,U}}[\Lscr^{\trte}_{g,t}(U)-\Lscr_t^{\trte}(U|\mset)]$, which represents the expected environment-level uncertainty measured using the average per-task training loss $L^{\trte}_{t}(u|Z^m)$ defined in \eqref{eq:modified_loss}. 
To this end,  we extend the single-task learning generalization bound of Lemma~\ref{cor:MI_singletask_subGaussian} by 
resorting to the decoupling estimate in Lemma~\ref{lem:decoupling} with $X=U$, $Y=\mset$ and $f(X,Y)=\Lscr^{\trte}_t(U|\mset)$, so that $\Ebb_{P_{X,Y}}[f(X,Y)]=\Ebb_{P_{\mset,U}}[\Lscr^{\trte}_t(U|\mset)]$ and $\Ebb_{P_XP_{Y}}[f(\Xt,\Yt)]= \Ebb_{P_{\mset,U}}[\Lm_{g}^{\trte}(U)]$.

Recall from Assumption~\ref{assum:2} that for a given $u \in \Uscr$, $\Lscr_t^{\trte}(u|\mset)=\frac{1}{N}\sum_{i=1}^N L_t^{\trte}(u|Z^m_i)$ is the average of $\sigma^2$-sub-Gaussian i.i.d. random variables $L_t^{\trte}(u|Z^m_i)$ under $Z^m_i \sim P_{Z^m}$ for all $i \in \{1,\hdots,N\}$. It then follows that $\Lscr_t^{\trte}(u|\mset)$ is $\sigma^2/N$-sub-Gaussian under $\mset \sim P_{\mset}$ for all $u \in \Uscr$ \citep{boucheron2013concentration}. 
Applying Lemma~\ref{lem:decoupling} with $\Psi^{*-1}_{\mp}$ specialized to the $\sigma^2/N$-sub-Gaussian loss in \eqref{eq:subgaussian-general} gives that
\begin{align}
 \biggl |\Ebb_{P_{\mset,U}} \biggl[\Lm^{\trte}_{g,t}(U)-\Lscr_t^{\trte}(U|\mset)\biggr] \biggr|
&\leq \sqrt{\frac{2 \sigma^2 I(U;\mset)}{N}}. \label{eq:boundingargument}
\end{align}

We now evaluate the first difference in \eqref{eq:auxiliary _metagap}. It can be seen that for a fixed task $\tau \in \mathcal{T}$, the average within-task uncertainty evaluates to
\begin{align}
\Ebb_{P_{Z^m|T=\tau}P_{W,U,\mset|\Ztrain}}[\DL(W|\Ztest,T=\tau)] 
&=\Ebb_{P_{Z^m|T=\tau}P_{W|\Ztrain}}[\DL(W|\Ztest,T=\tau)] \non \\&
= \Ebb_{P_{\Ztrain|T=\tau}}\Ebb_{P_{W|\Ztrain}}[L_{g}(W|T=\tau)-\Ebb_{P_{\Ztest|T=\tau}}L_{t}(W|\Ztest)] \non \\&
\stackrel{(a)}{=}
0, \label{eq:zeroloss}
\end{align}
where $(a)$ follows since $W$ and $\Ztest$ are conditionally independent given $\Ztrain$, whereby  $\Ebb_{P_{\Ztrain|T=\tau}}\Ebb_{P_{W|\Ztrain}}[\Ebb_{P_{\Ztest|T=\tau}}L_{t}(W|\Ztest)]=\Ebb_{P_{\Ztrain|T=\tau}}\Ebb_{P_{W|\Ztrain}}[L_{g}(W|T=\tau)]$.
Substituting  \eqref{eq:boundingargument} and \eqref{eq:zeroloss} in \eqref{eq:auxiliary _metagap} then concludes the proof.
For Theorem~\ref{thm:ITMI_separate}, the proof follows along the same line bounding the average environment-level uncertainty
$\Ebb_{P_{\mset,U}}[\Lscr^{\trte}_{g,t}(U)-\Lscr_t^{\trte}(U|\mset)]$. Towards this, we note that the environment-level uncertainty can be equivalently written as
\begin{align}
\Ebb_{P_{\mset,U}}[\Lscr^{\trte}_{g,t}(U)-\Lscr_t^{\trte}(U|\mset)]&=\frac{1}{N}\sum_{i=1}^N \biggl(\Ebb_{P_{Z^m_i} P_{U}}[L_{t}^{\trte}(U|Z^m_i)]-\Ebb_{P_{Z^m_i,U}}[L_{t}^{\trte}(U|Z^m_i)] \biggr), \label{eq:summation}
\end{align}
where $Z^m$ and $U$ in the first term are conditionally independent random variables distributed as $(Z^m_i,U)\sim P_{Z^m}P_{U}$, while, in the second term, they are jointly distributed according to $P_{Z^m_i,U}$, which is obtained by marginalizing the joint distribution $P_{\mset,U}$. 
Under Assumption~\ref{assum:bu_metaseparatesupport}$(a)$, we can bound the difference $\Ebb_{P_{Z^m_i} P_{U}}[L_{t}^{\trte}(U|Z^m_i)]-\Ebb_{P_{Z^m_i,U}}[L_{t}^{\trte}(U|Z^m_i)]$ by resorting to the decoupling estimate in Lemma~\ref{lem:decoupling} with $X=U,Y=Z^m_i$, $f(X,Y)=L_t^{\trte}(U|Z^m_i)$ such that $\Ebb_{P_{X,Y}}[f(X,Y)]=\Ebb_{P_{U,Z^m_i}}[L_{t}^{\trte}(U|Z^m_i)]$ and $\Ebb_{P_{X}P_{Y}}[f(\tilde{X},\tilde{Y})]=\Ebb_{P_{U}P_{Z^m_i}}[L_{t}^{\trte}(U|Z^m_i)]$. Since $L_t^{\trte}(U|Z^m_i)$ is $\sigma^2-$sub-Gaussian under Assumption~\ref{assum:bu_metaseparatesupport}$(a)$ for all $u \in \Uscr$, Lemma~\ref{lem:decoupling} yields the following bound
\begin{align}
\biggl | \Ebb_{P_{Z^m_i} P_{U}}[L_{t}^{\trte}(U|Z^m_i)]-\Ebb_{P_{Z^m_i,U}}[L_{t}^{\trte}(U|Z^m_i)] \biggr | \leq \sqrt{2 \sigma^2 I(U;Z^m_i)}. \label{eq:bound_1}
\end{align}

The bound in \eqref{eq:bound_1} can also be obtained using Assumption~\ref{assum:bu_metaseparatesupport}$(b)$ by resorting to the general decoupling estimate in Lemma~\ref{lem:general_decoupling} by fixing $X=U,Y=Z^m_i$, $f(X,Y)=L_t^{\trte}(U|Z^m_i)$ such that $\Ebb_{P_{X,Y}}[f(X,Y)]=\Ebb_{P_{U,Z^m_i}}[L_{t}^{\trte}(U|Z^m_i)]$ and $\Ebb_{P_{X}P_{Y}}[f(\tilde{X},\tilde{Y})]=\Ebb_{P_{U}P_{Z^m_i}}[L_{t}^{\trte}(U|Z^m_i)]$. Substituting the bound in \eqref{eq:bound_1} in \eqref{eq:summation} then yields the required bound in \eqref{eq:FMMI_separate_subGaussian}.
%
\section{Proofs of Theorem~\ref{thm:FMMIbound_same} and Theorem~\ref{thm:ITMIbound_same}}\label{app:1}
For Theorem~\ref{thm:FMMIbound_same}, we start from the following decomposition of the average meta-generalization gap analogous to \eqref{eq:auxiliary _metagap}
\begin{align}
&\Ebb_{P_{\mset,U}}[\Delta \Lscr^{\jnt}(U|\mset)] =\Ebb_{P_T} \biggl[\Ebb_{P_{Z^m|T}P_{W|Z^m}}[\DL(W|Z^m,T)]) \biggr]+\Ebb_{P_{\mset,U}}[\Lscr^{\jnt}_{g,t}(U)-\Lscr_t^{\jnt}(U|\mset)] \label{eq:auxiliary _metagap_2}
\end{align} 
where $
\Lm^{\jnt}_{g,t}(u)=\Ebb_{P_{T,Z^m}}[L^{\jnt}_t(u|Z^m)] 
$ is the average training loss for the randomly selected meta-test task as a function of the hyperparameter $u$, and $\DL(W|Z^m,T) =L_{P_{Z|T}}(W)-L_{t}(W|Z^m)$ is the generalization gap for the meta-test task $T$.
 The MI bound on 
 the expected environment-level uncertainty, $\Ebb_{P_{\mset,U}}[\Lscr^{\jnt}_{g,t}(U)-\Lscr_t^{\jnt}(U|\mset)]$, can be obtained by using Lemma~\ref{lem:decoupling} and the Assumption~\ref{assum:FMMIjoint}$(b)$ as in \eqref{eq:boundingargument}. 
 
 The main difference between the separate and joint within-task training and test sets scenarios is that while the average within-task uncertainty vanishes in the former scenario, this is not the case for joint within-task training and training sets.  Consequently, we now bound the average within-task generalization gap denoted by the first difference in \eqref{eq:auxiliary _metagap_2}.  For given task $\tau \in \mathcal{T}$, to bound the within-task generalization gap $\Ebb_{P_{Z^m|T=\tau}P_{W|Z^m}}[\DL(W|Z^m,T=\tau)]$, we resort to 
%
%
Lemma~\ref{lem:decoupling} with $X=W$, $Y=Z^m$ and $f(X,Y)=L_{t}(W|Z^m)$, so that
$\Ebb_{P_{X,Y}}[f(X,Y)]=\Ebb_{P_{W,Z^m|T=\tau}}[L_{t}(W|Z^m)]$. It can be then verified that $\Ebb_{P_X P_Y}[f(\Xt,\Yt)]=\Ebb_{P_{W|T=\tau}}\Ebb_{P_{Z^m|T=\tau}}[L_t(W|Z^m)]=\Ebb_{P_{W|T=\tau}}[L_g(W|T=\tau)]=\Ebb_{P_{W,Z^m|T=\tau}}[L_{g}(W|T=\tau)]$, where $
P_{W,Z^m|T=\tau}=P_{W|Z^m} P_{Z^m|T=\tau} $. Since $L_t(w|Z^m)$ is the sum of i.i.d $\delta_{\tau}^2$-sub-Gaussian random variables $l(w,Z_i)$ (from Assumption~\ref{assum:FMMIjoint}$(a)$),  we have that $L_t(w|Z^m)$ is $\delta_{\tau}^2/m$-sub-Gaussian under $Z^m \sim P_{Z^m|T=\tau}$ for all $w \in \Wscr$ \citep{boucheron2013concentration}. Consequently, Lemma~\ref{lem:decoupling} yields the following bound
\begin{align}
    \biggl | \Ebb_{P_{Z^m|T=\tau}P_{W|Z^m}}[\DL(W|Z^m,T=\tau)] \biggr | \leq \sqrt{\frac{2 \delta_{\tau}^2}{m}I(W;Z^m|T=\tau)}. \label{eq:bound_22}
\end{align}
Averaging with respect to $P_T$ on both sides of \eqref{eq:bound_22}, and combining with the bound on average environment-level uncertainty yields the required bound in \eqref{eq:MI_joint_subGaussian} via Jensen's inequality.

For Theorem~\ref{thm:ITMIbound_same}, the proof follows along the same line.
 The ITMI bound on the expected environment-level uncertainty can be obtained along the lines of \eqref{eq:bound_1}, using the assumption on $L_t^{\jnt}(u|Z^m)$ in either Assumption~\ref{assum:ITMIjoint}$(a)$ or  Assumption~\ref{assum:ITMIjoint}$(b)$. We now show that we can similarly bound the within-task uncertainty using the assumption on loss function $l(w,z)$ in either Assumption~\ref{assum:ITMIjoint}$(a)$ or Assumption~\ref{assum:ITMIjoint}$(b)$.
Towards this, for fixed task $\tau \in \mathcal{T}$, we write  the average within-task uncertainty equivalently as
\begin{align}
\Ebb_{P_{Z^m|T=\tau}P_{W|Z^m}}[\DL(W|Z^m,T=\tau)]=\frac{1}{m}\sum_{j=1}^m \biggl( \Ebb_{P_{W|T=\tau} P_{Z_j|T=\tau}}[l(W,Z_j)]-\Ebb_{P_{W,Z_j|T=\tau}}[l(W,Z_j)] \biggr),
\end{align}
where $W$ and $Z_j$ in the second term are jointly distributed according to $P_{W,Z_j|T=\tau}$, which is the marginal of the joint distribution $P_{W,Z^m|T=\tau}$. In contrast,
$W$ and $Z_j$ in the first term are  conditionally independent random variables distributed as $(W,Z_j)\sim P_{W|T=\tau} P_{Z_j|T=\tau}$ where $P_{W|T=\tau}$ is the marginal distribution of $P_{W,Z_j|T=\tau}$.
Now, fixing $X=W$, $Y=Z_j$ and $f(X,Y)=l(W,Z_j)$ so that  $\Ebb_{P_X P_Y}[f(\Xt,\Yt)]=\Ebb_{P_{W|T=\tau} P_{Z_j|T=\tau}}[l(W,Z_j)]$ and $\Ebb_{P_{X,Y}}[f(X,Y)]=\Ebb_{P_{W,Z_j|T=\tau}}[l(W,Z_j)]$ in Lemma~\ref{lem:decoupling} under the assumption on $l(w,z)$ in Assumption~\ref{assum:ITMIjoint}$(a)$, or in Lemma~\ref{lem:general_decoupling} under the assumption on $l(w,z)$ in Assumption~\ref{assum:ITMIjoint}$(b)$ yields the following bound,
\begin{align}
\biggl | \Ebb_{P_{Z^m|T=\tau}P_{W|Z^m}}[\DL(W|Z^m,T=\tau)]\biggr | \leq \frac{1}{m}\sum_{j=1}^m \sqrt{2 \delta_{\tau}^2 I(W;Z_j|T=\tau)} \label{eq:bound_3}.
\end{align} Averaging with respect to $P_T$ on both sides of \eqref{eq:bound_3}, and combining with the bound on average environment-level uncertainty yields the required bound in \eqref{eq:ITMI_joint_subGaussian}.
\section{Details of Example}\label{app:example}
We first give details of the derivation of meta-generalization gap for the case with separate within-task training and test sets.
The average meta-generalization loss can be computed as $\Ebb_{P_{\mset,U}}[\Lm^{\trte}_{g}(U)]=$
\begin{align}
&\Ebb_{P_{\mset,U}}\biggl[(1-\alpha)^2U^2+\Ebb_{P_{T,\Ztrain}}\bigl[\alpha^2 (\Ztr_T)^2+\Ebb_{P_{Z|T}}[Z^2]-2\alpha\Ztr_T \mu_T +2 (1-\alpha)U(\alpha \Ztr_T-\mu_T) \bigr]\biggr]\non \\
&\stackrel{(a)}{=}\Ebb_{P_{\mset,U}}\biggl[(1-\alpha)^2 \bigl(U^2-2U \Ebb_{P_T}[\mu_T]\bigr)\biggr]+\Ebb_{P_{T}}\biggl[\alpha^2 \biggl(\mu_T^2+\frac{\mu_T \bar{\mu}_T}{\mtr}\biggr)+\mu_T-2\alpha\mu_T^2 \biggr], \label{eq:example_avgtest_sep}
\end{align}where the equality in $(a)$ follows since $\Ebb_{P_{Z|T}}[Z^2]=\mu_T$, $\Ebb_{P_{\Ztrain|T}}[\Ztr_T]=\mu_T$ and $\Ebb_{P_{\Ztrain|T}}[(\Ztr_T)^2]= \mu_T^2+\mu_T \bar{\mu}_T/\mtr$.
 In a similar manner,  the average meta-training loss can be computed as
\begin{align}
\Ebb_{P_{\mset,U}}[\Lscr^{\trte}_{t}(U|\mset)] &= \Ebb_{P_{\mset}}\biggl[ -(1-\alpha)^2U^2+\frac{1}{N}\sum_{i=1}^N\alpha^2(\Ztr_i)^2 \non \\&+\frac{1}{N}\sum_{i=1}^N \frac{1}{\mte}\sum_{j=1}^{\mte}(Z^{\mte}_{i,j})^2-2\alpha \frac{1}{N}\sum_{i=1}^N \Ztr_i \Zte_i\biggr],\label{eq:example_avgtrain_sep}
\end{align} with $U$ defined as in \eqref{eq:theta_separate}.
The meta-generalization gap in \eqref{eq:example_metagap_separate} then results by taking the difference of \eqref{eq:example_avgtest_sep} and \eqref{eq:example_avgtrain_sep}, and using that $\Ebb_{P_{\mset}}\bigl[(1-\alpha)^2 U^2 \bigr]=$
$
\Ebb_{P_T}[\mu_T \bar{\mu}_T]\bigl(\frac{1}{N\mte}+\frac{\alpha^2}{N\mtr}\bigr)+\frac{1}{N}(1+\alpha^2){\rm Var}_T+(1-\alpha)^2 (\Ebb_{P_T}[\mu_T])^2 
$ and $\Ebb_{P_{\mset}}[U]=\Ebb_{P_T}[\mu_T]$ with ${\rm Var}_T=\Ebb_{P_T}[\mu_T^2]-(\Ebb_{P_T}[\mu_T])^2 $.

We now evaluate the mutual informations $I(U;\mset)$ and $I(U;Z^m_i)$.
For the first MI, note that since the meta-learner is deterministic (see \eqref{eq:theta_separate}), $H(U|\mset)=0$ and thus $I(U;\mset)=H(U)$.
For the second MI, we can write $I(U;Z^m_i)=H(U)-\Ebb_{Z^m_i}[H(U|Z^m_i=z^m)]$.
It can be seen that random variables $U$ and $U|Z^m_i=z^m$ are mixtures of probability distributions, whose entropies can be evaluated following standard methods \citep{michalowicz2013handbook}.

For the case with joint within-task training and test sets, 
 the meta-generalization gap can be obtained in a similar way as 
\begin{align}
\Ebb_{P_{\mset,U}}[\Lscr^{\jnt}_{t}(U|\mset)]=\frac{2}{N}\biggl[ \Ebb_{P_T}\biggl[ \frac{\mu_T \bar{\mu}_T}{m}\biggr]+{\rm Var}_T\biggr]+ \frac{2\alpha}{m} \Ebb_{P_T}[\mu_T \bar{\mu}_T]
 \label{eq:example_metagap_joint}.
\end{align}

For the MI and ITMI-based bounds, note that with $\Wscr=[0,1]$, the loss function $l(\cdot,\cdot)$ is $[0,1]$-bounded, and for the deterministic base-learner in \eqref{eq:example_phi} with $\Uscr=[0,1]$, the average training loss $L_{t}^{\jnt}(u|Z^m)$ is also $[0,1]$-bounded  for all $Z^m \in \Zscr^m$.
  Thus, Assumption~\ref{assum:FMMIjoint} and Assumption~\ref{assum:ITMIjoint} hold with $\sigma^2=\delta_{\tau}^2=1/4$.. 
For the MI bound in \eqref{eq:MI_joint_subGaussian}, we have $I(U;\mset)=H(U)$ and 
 $I(W;Z^m|T=\tau)=H(W|T=\tau)-\Ebb_{Z^m}[H(W|Z^m,T=\tau)]$.
For the ITMI bound \eqref{eq:ITMI_joint_subGaussian}, we have
\begin{align}
|\Ebb_{P_{\mset,U}}[\Lscr^{\jnt}_{t}(U|\mset)]|& \leq \frac{1}{N}\sum_{i=1}^N\sqrt{\frac{1}{2}I(U;Z^m_i)}+\Ebb_{P_T}\biggl[ \frac{1}{m}\sum_{j=1}^m \sqrt{\frac{1 }{2}I(W;Z_j|T=\tau)}\biggr]. \label{eq:exampleITMI_joint}
\end{align}
All information measures can be easily evaluated numerically \citep{michalowicz2013handbook}.
\section{Proof of Lemma~\ref{lem:noisyalgorithm}}\label{app:iterative}
 From the update rule of the meta-learner in \eqref{eq:updaterule}, we get the Markov dependency
\begin{align}
&P_{U^j|U^{(j-1)},\{W_{K_i}\}_{i=1}^{j}, \{Z^m_{K_i}\}_{i=1}^j,\mset}=P_{U^j|U^{j-1},W_{K_j}, \Ztest_{K_j}}, \label{eq:Markov_metalearner}
\end{align}
where $U^{(j-1)}=\{U^1,\hdots,U^{j-1}\}$ is the history vector of hyperparameters.
The sampling strategy in \eqref{eq:sampling} together with \eqref{eq:Markov_metalearner}  then implies the following relation
\begin{align}
P_{U^j|U^{(j-1)},\{W_{K_i}\}_{i=1}^{j}, \{Z^m_{K_i}\}_{i=1}^J,\mset} =P_{U^j|U^{j-1},W_{K_j}, \Ztest_{K_j}}. \label{eq:sampling+update}
\end{align}

Using $U^{(J)}=\{U^1,\hdots,U^J\}$ to denote the set of all updates, we have the following  relations
\begin{align}
I(U;\mset)&\stackrel{(a)}{\leq } I(U^{(J)};\mset)\non \\
&\stackrel{(b)}{\leq}I(U^{(J)};\{Z^m_{K_i}\}_{i=1}^J)
=\sum_{j=1}^J I(U^j;\{Z^m_{K_i}\}_{i=1}^J|U^{(j-1)})\\
&\leq \sum_{j=1}^J I(U^j;\{Z^m_{K_i}\}_{i=1}^J,\{W_{K_i}\}_{i=1}^j|U^{(j-1)})\\
&=\sum_{j=1}^J h(U^j|U^{(j-1)})-h\biggl(U^j|U^{(j-1)},\{Z^m_{K_i}\}_{i=1}^J,\{W_{K_i}\}_{i=1}^j \biggr)\\
&\stackrel{(c)}{=} \sum_{j=1}^J\biggl[h(U^j|U^{j-1})-h(U^j|U^{j-1},W_{K_j},\Ztest_{K_j}) \biggr],
 \label{eq:boundcomputation}
\end{align}
where, the inequality in $(a)$ follows from data processing inequality on Markov chain $\mset \rightarrow U^{(J)} \rightarrow U$; $(b)$ follows from the Markov chain $\mset \rightarrow \{Z^m_{K_i}\}_{i=1}^J \rightarrow U^{(J)}$; and  the equality in $(c)$ follows from $U^{(j-2)} \rightarrow U^{j-1} \rightarrow U^j$ and \eqref{eq:sampling+update}.
Finally, the computation of bound in \eqref{eq:boundcomputation} follows similar to Lemma~5 in \citep{pensia2018generalization}. 
\externalbibliography{yes}
\bibliography{ref}



\end{document}